\documentclass[journal]{IEEEtran} %Ë«À¸
\usepackage{textcomp}
\usepackage{booktabs}
\usepackage{colortbl}
\usepackage[table]{xcolor}
\usepackage{graphicx}
\usepackage{subfigure}
\usepackage{multirow}
\usepackage{longtable}
\usepackage{algorithm}
\usepackage{algorithmic}
\usepackage{booktabs}
\usepackage{stfloats}
\usepackage{caption2}
\usepackage{amssymb,amsmath}
\usepackage{epstopdf}
\usepackage{color}
\usepackage{rotating}
\usepackage{multirow}
\usepackage{threeparttable}
\usepackage{bbding}
\usepackage{tabularx}
\usepackage{booktabs}
\usepackage{cite}
\usepackage[colorlinks,linkcolor=red,anchorcolor=blue,citecolor=blue]{hyperref}

\usepackage{orcidlink}

\definecolor{rrr}{rgb}{1    0	0.0392}
\definecolor{myred}{rgb}{0.7059    0	0.0392}					% revised contents for reviewer 1
\definecolor{myblue}{rgb}{0    0	0.9392}					% revised contents for reviewer 1
\ifCLASSINFOpdf
\else
\fi
\begin{document}
\title{ERDDCI: Exact Reversible Diffusion via Dual-Chain Inversion for High-Quality Image Editing}
%\title{Co-Optimization: Learning Multi-source Perception Representation for Camouflaged Object Detection}
\author{
        Jimin~Dai, 
        Yingzhen~Zhang,
	Shuo~Chen, 
        Jian~Yang, %\emph{Member,IEEE},
        Lei~Luo %\textsuperscript{\Envelope}

\thanks{This work was supported by the National Natural Science Foundation of China (Grant No. 62276135, 61806094), %Natural Science Research Project of Colleges and Universities in Anhui Province (Grant No. xxxxxxx). 
(\textit{Corresponding author: Lei Luo,} e-mail: luoleipitt@gmail.com)
	
J. Dai, Y. Zhang, L. Luo, and J. Yang are with the PCA Laboratory, Key Laboratory of Intelligent Perception and Systems for High-Dimensional Information of Ministry of Education, School of Computer Science and Engineering, Nanjing University of Science and Technology, Nanjing, China. (e-mail: jimindai@njust.edu.cn; yingzhenzhang@njust.edu.cn; luoleipitt@gmail.com; csjyang@mail.njust.edu.cn).

S. Chen is with the RIKEN, Japan. (shuo.chen.ya@riken.jp)

%J. Yang is with the College of Computer Science and Engineering, Nankai University, Tianjin, 300000, China. (csjyang@nankai.edu.cn)
}
}

\markboth{Journal of \LaTeX\ Class Files}%
{Shell \MakeLowercase{\textit{et al.}}: Bare Demo of IEEEtran.cls for IEEE Journals}

\maketitle
\begin{abstract}
Diffusion models (DMs) have been successfully applied to real image editing. These models typically invert images into latent noise vectors used to reconstruct the original images (known as inversion), and then edit them during the inference process. However, recent popular DMs often rely on the assumption of local linearization, where the noise injected during the inversion process is expected to approximate the noise removed during the inference process. 
While DM efficiently generates images under this assumption, it can also accumulate errors during the diffusion process due to the assumption, ultimately negatively impacting the quality of real image reconstruction and editing. To address this issue, we propose a novel method, referred to as ERDDCI (Exact Reversible Diffusion via Dual-Chain Inversion). ERDDCI uses the new Dual-Chain Inversion (DCI)  for joint inference to derive an exact reversible diffusion process. By using DCI, our method effectively avoids the cumbersome optimization process in existing inversion approaches and 
achieves high-quality image editing. Additionally, to accommodate image operations under high guidance scales, we introduce a dynamic control strategy that enables more refined image reconstruction and editing. Our experiments demonstrate that ERDDCI significantly outperforms state-of-the-art methods in a 50-step diffusion process. It achieves rapid and precise image reconstruction with an SSIM of 0.999 and an LPIPS of 0.001, and also delivers competitive results in image editing.

\end{abstract}
\begin{IEEEkeywords}
Diffusion models, image editing,  dull-chain inversion. 
\end{IEEEkeywords}
\IEEEpeerreviewmaketitle

\section{INTRODUCTION}
\IEEEPARstart{D}iffusion models (DMs) are a type of probabilistic generative models that transform data into noise through a forward diffusion process, followed by a reverse generative process that iteratively denoises to produce meaningful data \cite{sohl2015deep,ho2020denoising,song2020denoising,saharia2022palette,ho2022classifier,peebles2023scalable}. 
The rapid advancement of this technology has significantly enhanced the quality and diversity of image generation and editing \cite{chen2021multi,xu2023txt2img,schwartz2018deepisp,yang2023eliminating,zhao2021efficient}. 
Models such as Stable Diffusion \cite{rombach2022high}, DALL-E2 \cite{ramesh2022hierarchical}, and GLIDE \cite{nichol2021glide} are capable of generating high-quality images from text and support controllable editing based on the generated images. 
However, due to DMs' inherent design logic, which involves step-by-step inference to invert sampled Gaussian noise into images \cite{gilboa2002forward}, editing real images remains a challenge. This difficulty arises because DMs must first convert real images into their corresponding Gaussian noise, a process referred to as “inversion” \cite{zhang2024real}, before effectively performing image editing tasks during the inference process.
\begin{figure*}[t!]
\centering
\includegraphics[width=\textwidth]{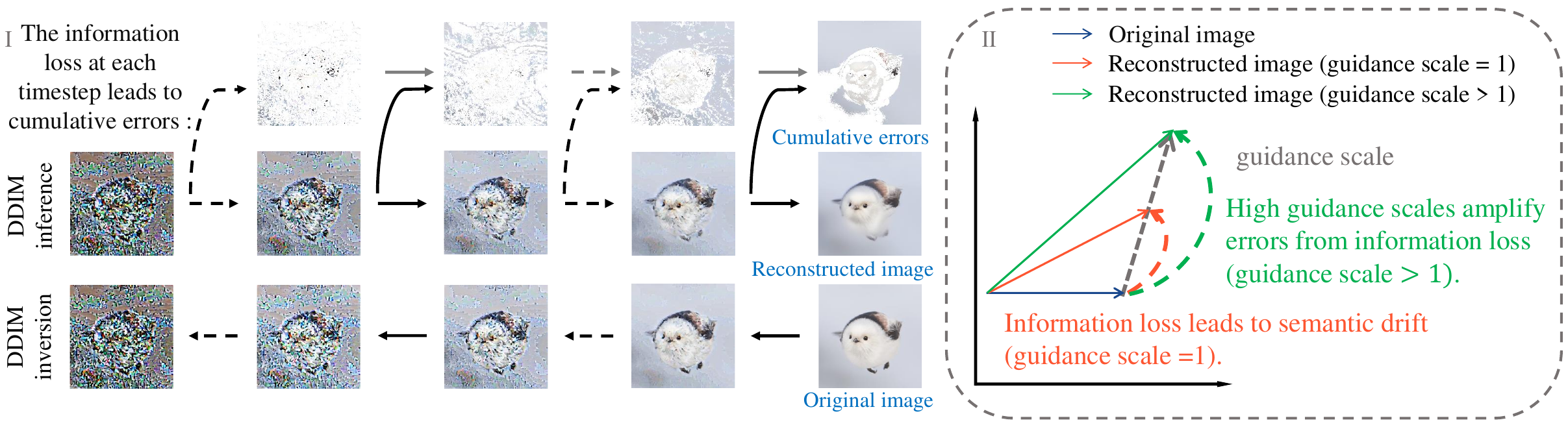} % Reduce the figure size so that it is slightly narrower than the column.
\caption{{\bf Error Accumulation and Amplification in DDIM Inference.} The DDIM adheres to a local linearization assumption. This assumption accumulates errors step by step during the inference process following image inversion (Part \MakeUppercase{\romannumeral 1}), causing the semantic information of the reconstructed image to deviate from the original image. These errors are further amplified under high guidance scales (Part \MakeUppercase{\romannumeral 2}), affecting the fidelity of the image. Black arrows represent the inference or inversion process. Black curved arrows indicate that each step generates errors. Gray arrows demonstrate the gradual accumulation of errors. Solid arrows signify single-step processes, and dashed arrows denote multi-step processes.}
\label{fig:error}
\end{figure*}

High-quality edited images should satisfy the following criteria: (i) The edited image should maintain high fidelity with the original, meaning it should not alter the original layout or make unnecessary semantic changes beyond the editing requirements. (ii) The edited image must meet the editing requirements, specifically, it must accurately modify the semantic information specified in the editing requirements. 
As both image reconstruction and editing are conducted through the DM's inference process, with editing introducing additional conditions during inference. Therefore, the prerequisite for high-quality real image editing is precise image reconstruction, which means restoring the original image from the noise obtained through inversion. Precise reconstruction can ensure that the edited image remains high fidelity with the original image. The first step in image reconstruction is obtaining the noise from inversion, many real image editing models (such as Prompt-to-Prompt (PTP) \cite{hertz2022prompt}) use Denoising Diffusion Implicit Models (DDIM) \cite{song2020denoising} for inversion, as shown in the third row of part \MakeUppercase{\romannumeral 1} in Fig.\ref{fig:error}, where real images are progressively inverted to noise. After obtaining the noise corresponding to the real image, DDIM's inference process can gradually reconstruct the image, as illustrated in the second row of part \MakeUppercase{\romannumeral 1} in Fig.\ref{fig:error}. 

However, due to the local linearization assumption \cite{wallace2023edict,pan2023effective} followed by DDIM, along with model training errors and rounding errors from underlying computations, partial information is lost during the diffusion process, leading to gradually accumulating errors, as shown in the first row of part \MakeUppercase{\romannumeral 1} in Fig.\ref{fig:error}. These accumulated errors cause the semantic information of the reconstructed image to deviate from the original image, further impacting the quality of edited images. Additionally, under high guidance scales, these accumulated errors further increase \cite{ho2022classifier,han2024proxedit}. As illustrated in part \MakeUppercase{\romannumeral 2} of Fig.\ref{fig:error}, the green arrow deviates from the blue arrow by a larger angle compared to the red arrow. This indicates that as the guidance scale increases, the semantic deviation between the reconstructed images and the original images grows larger.

In this paper, we propose a new method, Exact Reversible Diffusion via Dual-Chain Inversion (ERDDCI), to eliminate these accumulated errors caused by the linearization assumption under classifier-free guidance. Our approach involves Dual-Chain Inversion (DCI) technology with a dynamic editing strategy, resulting in improved quality and precision for real image reconstruction and editing. The inception of the DCI is developed through a thorough analysis of the DM's process, aiming to address the persistent issue of the local linearization assumption that previous research has been unable to overcome \cite{liu2022pseudo}. By incorporating a new auxiliary inversion chain with the original DDIM inversion chain for joint inference, we circumvented the errors caused by local linearization assumptions in diffusion mechanism. Our method has demonstrated effectiveness through experimental validations, surpassing state-of-the-art methods in this domain. Unlike recent approaches \cite{mokady2023null,pan2023effective},  ERDDCI avoids complex optimization procedures in favor of a more streamlined and mathematically grounded approach. This not only improves the precision and reversibility of the diffusion process but also significantly reduces computational demands. 
As shown in Tab.\ref{tab:duration}, our method requires much less time than other optimized inversion methods. 

Additionally, we introduce a Dynamic Control Strategy (DCS) to further refine image reconstruction and editing. In recent research, PTP offers a user-friendly image editing method without needing extra guidance inputs, such as masks. However, using classifier-free guidance with high scales in PTP can severely distort the semantic information of the original image, resulting in the edited image differs significantly from the original image.
Null-Text inversion (NTI) \cite{mokady2023null} achieves better editing outcomes at default guidance scales ($\omega $=7.5) by adding optimization during the inference process to minimize the semantic distance between the edited and original images, offsetting the effects of high guidance scales. 
This approach implicitly constrains the influence of the guiding scales on the image during the editing process. 
In contrast, DCS aims to explicitly and intuitively constrain the guidance scales.
Our observations, along with related research \cite{biroli2024dynamical}, show that during the denoising phase, noise typically %generally  
first outlines the image's broad layout, then adds detailed information based on this layout, and finally performs fine-grained generation in the last stage of denoising to achieve %for
the desired visual effect (as shown in the second row of Part \MakeUppercase{\romannumeral 1} in Fig.\ref{fig:error}). 
Leveraging this, we employ a dynamic control of guidance scales for image editing, adjusting the guidance scale throughout the editing process for more precise edits. 
Our experiments demonstrate the effectiveness of DCS.

In summary, based on the pre-trained Stable DM, we propose an innovative ERDDCI:
\begin{itemize}
  \item[•] By using DCI, we eliminate the adverse effects of accumulated errors during the inference process, ensuring high fidelity between the reconstructed/edited images and the original images.
  \item[•]Implementing DCS allows for more precise and controllable image editing at high guidance scales, leading to superior outcomes.
  \item[•]ERDDCI ircumvents intricate optimization procedures, significantly reducing computational costs and improving temporal efficiency compared to traditional methods.
\end{itemize}

\section{Related Work}

\subsection{Inversion in Diffusion Models}
Editing real images \cite{han2024proxedit,pehlivan2023styleres, brack2024ledits++} requires inversion to obtain the corresponding latent noise vector, followed by denoising processing and editing during the generative process.
denoising diffusion probabilistic models (DDPM) \cite{ho2020denoising}, a seminal work in DMs, led to CycleDiffusion's \cite{wu2022unifying} attempt at image editing via DDPM inversion. 
However, its native noise space limits editing capabilities.
Inbar et al. \cite{huberman2023edit} enhanced editability by creating a more editable DDPM noise space through independent Gaussian noise injection during DDPM inversion. 
DDIM breaks the Markov chain structure assumption of DDPM and significantly speeds up DM generation. 
Currently, mainstream inversion uses DDIM, but its linear assumption accumulates errors, destabilizing editing in conditional DMs, especially at high guidance scales as it exacerbates these errors.

To address these issues, modifications to the DDIM inversion have been explored \cite{zhang2025exact,tang2024iterinv,huberman2024edit}.
Accelerated Iterative Diffusion Inversion (AIDI) \cite{pan2023effective} focuses on the DDIM inversion process, employing fixed-point iteration to optimize each step's output, which then serves as the input for subsequent inversions, thereby iteratively finding the optimal latent noise vector.
NTI optimizes the null text embedding during the generation process without directly altering the DDIM inversion. It references the inversion trajectory, a sequence of latent noise vectors computed by the inversion process, to keep the edited image semantically similar to the original.
Following a similar approach, Negative-Prompt Inversion (NPI) \cite{miyake2023negative} uses the original image's prompt to replace null text embedding for optimization in the generation process, yielding favorable results.
Similarly, Stylediffusion \cite{li2023stylediffusion} uses the DDIM inversion trajectory as a pivot, optimizing attention maps and latent image variables at each step in the generation process to gain effective editing outcomes.
Exact Diffusion Inversion via Coupled Transformations (EDICT) \cite{wallace2023edict} operates simultaneously on both inversion and generation processes, implementing precise inversion through symmetrically coupled transformations, thereby achieving effective image reconstruction and editing. However, these inversion methods are resource-intensive and time-consuming, failing to strike an ideal balance between high computational efficiency and precision in image operation. 
In contrast, our method achieves lightweight, rapid, and high-quality image operations.
\subsection{Real Image Editing in Diffusion Models}
Editing real images encompasses various technical approaches.
SDEdit \cite{meng2021sdedit}, Blended Diffusion \cite{avrahami2022blended}, and DiffEdit \cite{couairon2022diffedit} allow for localized editing of real images through stroke or mask guidance coupled with random noise perturbations.
DreamBooth \cite{ruiz2023dreambooth} fine-tunes pre-trained DMs to bind subjects to unique identifiers, embedding multiple subject instances in the model's output domain for semantic editing of the subject's actions, environments, and so on. 
Advancing further, Imagic \cite{kawar2023imagic} aligns text embeddings with a single input image and target text, fine-tuning the DMs to edit the subject of the input image. 
For global image editing, DDIB \cite{su2022dual} employs DDIM to incrementally introduce deterministic noise, in tandem with fine-tuning pre-trained models to achieve desired outcomes.
DiffusionCLIP \cite{kim2022diffusionclip} fine-tunes the scoring function in the reverse diffusion process using CLIP loss, tailoring edited image attributes to textual prompts.

To circumvent the extensive computational demand of fine-tuning, BoundaryDiffusion \cite{zhu2023boundary}, a non-learning approach, operates directly in latent space with boundary-guided hybrid trajectories for effective semantic image editing. 
Another innovative method, such as Text2LIVE \cite{bar2022text2live}, does not operate directly on the original image but creates an editing layer for localized image editing, thereby maintaining the integrity of the other parts of the picture without changes.

Additional real image editing methods involve the ideas of PTP and the application of inversion. 
For instance, DiffEdit generates masks for editing areas based on source and target texts, ensuring fidelity to both unedited regions to the original image and edited areas to the target text. 
NTI stabilizes editing at higher guidance scales by refining the inference process. 
AIDI introduces blended guidance and applies high guidance scales to the edited sections while setting the guidance scale to 1 for the unedited parts, thereby ensuring fidelity to the original image. Some of the editing strategies mentioned above have significant limitations or high learning costs and fail to meet user requirements. Therefore, we propose DCS to achieve more refined and controllable image editing. 

\section{Methodology}
\subsection{Preliminaries}
\label{sec:3.1}
The text-guided implicit DM translates the original diffusion process from pixel to latent space.
Given an image $x_{0}$, it's first encoded as $z_{0}=Enc(x_{0})$, and then subjected to a forward diffusion process over $T$ steps, transforming $z_{0}$ into pure noise $z_{T}$ through the addition of noise:
\begin{equation}
  z_{t}=\sqrt{\alpha _{t} }  z_{t-1} +\sqrt{1-\alpha  _{t} } \epsilon , ~~~ \epsilon \sim N(0,I),~~~t=1,2,\dots ,T ,
  \label{eq:1}
\end{equation}
because of the linear Gaussian assumption, we can calculate any $z_{t}$ from $z_{0}$ in a single step:
\begin{equation}
  z_{t}=\sqrt{\bar{\alpha }_{t} }  z_{0} +\sqrt{1-\bar{\alpha}  _{t} } \epsilon , ~~~ \epsilon \sim N(0,I),~~~t=1,2,\dots ,T ,
  \label{eq:2}
\end{equation}
where $\epsilon$ is sampled from standard Gaussian noise, $\bar{\alpha}  _{t} : =  {\textstyle \prod_{i=1}^{t}} \alpha _{i} $, $\left \{  \alpha _{t}  \right \}_{t=1 }^{T} $ is a manually set schedule.

During the backward diffusion process, a trained noise prediction network $\epsilon _{\theta } $ is utilized to denoise $z'_{t}$ to generate $z'_{t-1}$, where $z'_{T}=z_{T}$. With the DDIM sampler, it can be expressed as:
\begin{equation}
 z'_{t-1}=\sqrt{ \bar{\alpha}_{t-1} }  f_{\theta  } (z'_{t})+\sqrt{1-\bar{\alpha}_{t-1}} \Tilde{\epsilon}(z'_{t},t).
  \label{eq:3}
\end{equation}

Here, $ f_{\theta  } (z'_{t})$ represents the prediction of $z_{0}$ at timestep $t$, following the local linearization assumption:
\begin{equation}
   f_{\theta  } (z'_{t})=\frac{z'_{t}-\sqrt{1-\bar{\alpha}_{t}} \tilde{\epsilon}\ }{\sqrt{\bar{\alpha} _{t} } }  ,
  \label{eq:4}
\end{equation}
and $\Tilde{\epsilon}$ is a noise prediction network operating under classifier-free guidance, denoted as:
\begin{equation}
  \Tilde{\epsilon}(z'_{t},t)=\omega   \epsilon _{\theta }(z'_{t},C,t)+(1-\omega )\epsilon _{\theta }(z'_{t},\emptyset ,t),
  \label{eq:5}
\end{equation}
where $\omega$ is the classifier-free guidance scale, $C$ is the conditional text embedding, $\emptyset$ is the unconditional text embedding, and $\epsilon _{\theta }$ is optimized using the following loss function:
\begin{equation}
 Loss=\mathbb{E}_{\epsilon ,z,C,t}\left \| \epsilon -\epsilon _{\theta }(z_{t},C,t) \right \| ^{2}_{2}.
  \label{eq:6}
\end{equation}

After the full backward diffusion process, the denoised latent image variable $z'_{0}$ is obtained, which is then decoded to yield $x'_{0}=Dec(z'_{0})$, theoretically, $x'_{0}$ should be identical to $x_{0}$.

For the DDIM inversion process, just replicate the forward diffusion process but replace the $\epsilon$ in each timestep with $ \Tilde{\epsilon}$ predicted by pre-trained noise prediction network:
\begin{equation}
 \hat{z}_{t}=\sqrt{ \bar{\alpha}_{t} }  f_{\theta  } (\hat{z} _{t-1})+\sqrt{1-\bar{\alpha}_{t}} \tilde{\epsilon}(\hat{z}_{t-1},t),
  \label{eq:7}
\end{equation}
where $\hat{z}_{t}$  is the latent image variable obtained at the $t$-th timestep of the inversion process, $\hat{z}_{0}=z_{0}$. 

\begin{figure*}
  \centering
  \includegraphics[width=\textwidth]{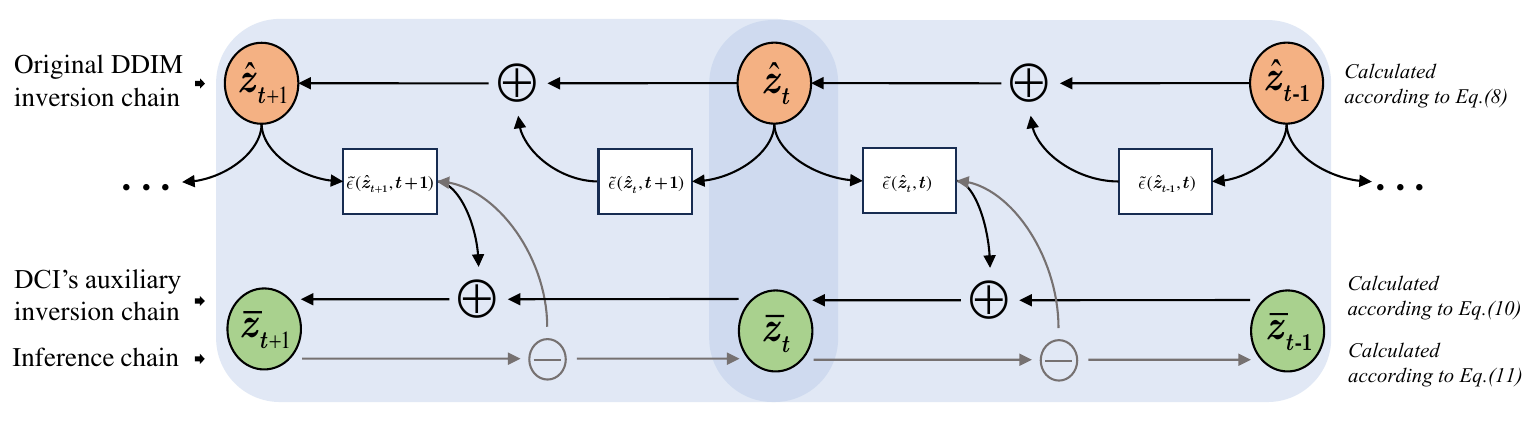}
  \caption{{\bf Dual-chain inversion overview.}   DCI makes DMs exactly reversible between the auxiliary inversion chain and the inference chain.  $\oplus$ represents the noise adding operation, which follows Eq.(\ref{eq:8}) in the original DDIM inversion chain, and follows Eq.(\ref{eq:10}) in the auxiliary inversion chain; $\ominus$ represents the denoising operation, following Eq.(\ref{eq:11}). 
  }
  \label{fig:overview}
\end{figure*}

In the above discourse, discrepancies occur among latent image variables at the same timestep $t$ across different processes due to errors instigated by local linearization assumptions. 
To facilitate the distinction, $z_{t}$ denotes the latent image variables in the forward diffusion process, $z'_{t}$ signifies those within the inference process, and $\hat{z}_{t}$ encapsulates those within the inversion process. 
DCI unifies the latent image variables from the inference and inversion processes, making these two processes completely reversible, thereby improving the fidelity and precision of the diffusion-based image reconstruction and editing processes. 
\subsection{Dual-Chain Inversion for Exact Reversible Diffusion}
As illustrated in Preliminaries, during the DDIM inversion process, $\hat{z}_{t-1}$ is transformed into $\hat{z}_{t}$ by predicting its noise $ \Tilde{\epsilon}(\hat{z}_{t-1},t)$ at timestep $t$ and injecting the noise into $\hat{z}_{t-1}$:
\begin{equation}
 \hat{z}_{t}=\sqrt{ \bar{\alpha}_{t} } \frac{\hat{z}_{t-1}-\sqrt{1-\bar{\alpha}_{t-1}} \tilde{\epsilon}(\hat{z}_{t-1},t)\ }{\sqrt{\bar{\alpha} _{t-1} } }+ \sqrt{1-\bar{\alpha}_{t}} \tilde{\epsilon}(\hat{z}_{t-1},t).
  \label{eq:8}
\end{equation}
However, in the inference process, $z'_{t}$ reverts to $z'_{t-1}$ by predicting its noise $\tilde{\epsilon}(z'_{t},t)$ at time $t$ and removing this noise:
\begin{equation}
 z'_{t-1}=\sqrt{ \bar{\alpha}_{t-1} } \frac{z'_{t}-\sqrt{1-\bar{\alpha}_{t}} \tilde{\epsilon}(z'_{t},t)\ }{\sqrt{\bar{\alpha} _{t} } }+ \sqrt{1-\bar{\alpha}_{t-1}} \tilde{\epsilon}(z'_{t},t).
  \label{eq:9}
\end{equation}
In both the inversion and inference processes, the noise injected and removed at the same timestep is not identical. 
DDIM operates under the local linearization assumption, approximating $\tilde{\epsilon}(\hat{z}_{t-1},t)$ to $\tilde{\epsilon}(z'_{t},t)$ to achieve an approximate reversible diffusion process,  formally expressed as $z'_{t} \approx \hat{z}_{t}$. 
%However, for more precise outcomes in practical applications, such as real image editing with classifier-free guidance, the error caused by this approximation cannot be overlooked.
However, this linearization can cause cumulative errors during the inference process, leading to deviations in the semantic information of the reconstructed images from the originals (as shown in the left part of Fig.\ref{fig:error}). 
Under high guidance scales, these semantic deviations are further amplified, significantly reducing the fidelity of edited images to their originals (as shown in the right part of Fig.\ref{fig:error}). 
Thus, eliminating these errors is crucial to ensuring precise reconstruction of the input images and improving the quality of image edits.
%for more precise outcomes in practical applications, such as real image editing with classifier-free guidance, the error caused by this approximation cannot be overlooked.
The DCI introduced in this paper aims to eliminate these errors %resulting from the local linearization assumption 
by incorporating another auxiliary inversion chain to do joint inference, thus ensuring the precision of the diffusion process for real-world applications.

We aim to ensure the precise alignment of injected and removed noise during both the inversion and inference processes, thereby achieving an exact reversible diffusion process. 
As depicted in Fig.\ref{fig:overview}, to attain this goal, in each step of the inversion process, we initially predict the noise $\tilde{\epsilon}(\hat{z}_{t-1},t)$ based on the current latent image variable $\hat{z}_{t-1}$ and the timestep $t$. 
This noise is then injected into $\hat{z}_{t-1}$ to obtain the latent image variable for the next timestep, $\hat{z}_{t}$, as shown in Eq.(\ref{eq:8}). 
Subsequently, to unify the inversion and inference phases, we calculate the noise $\tilde{\epsilon}(\hat{z}_{t},t)$, and then inject it into $\bar{z}_{t-1}$, which obtained through the same procedures as above, to achieve $\bar{z}_{t}$, \textit{i.e.},
\begin{equation}
\bar{z}_{t}=\sqrt{ \bar{\alpha}_{t} } \frac{\bar{z}_{t-1}-\sqrt{1-\bar{\alpha}_{t-1}} \tilde{\epsilon}(\hat{z}_{t},t)\ }{\sqrt{\bar{\alpha} _{t-1} } }+ \sqrt{1-\bar{\alpha}_{t}} \tilde{\epsilon}(\hat{z}_{t},t),
  \label{eq:10}
\end{equation}
where $\bar{z}_{0}=z_0$.
This operation is pivotal as it forms the auxiliary inversion chain, ensuring that the noise $\tilde{\epsilon}(\hat{z}_{t},t)$ injected during inversion at timestep $t$ would be precisely removed during the inference at timestep $t$.
% After completing the full inversion process, we acquire two inversion chains: $\left \{ \hat{z} _{t} \right \} ^{T}_{t=1}$ and $\left \{ \bar{z} _{t} \right \} ^{T}_{t=1}$, where $\left \{ \hat{z} _{t} \right \} ^{T}_{t=1}$ represents the inversion chain generated by the original DDIM inversion, and $\left \{ \bar{z} _{t} \right \} ^{T}_{t=1}$ is the auxiliary inversion chain obtained by additionally calculating the noise predicted for the next latent image variable at current timestep of the inversion process, and then reinjecting this noise into the latent image variable of the current timestep.
After completing the full inversion process, we acquire two inversion chains: $\left \{ \hat{z} _{t} \right \} ^{T}_{t=1}$ and $\left \{ \bar{z} _{t} \right \} ^{T}_{t=1}$, where $\left \{ \hat{z} _{t} \right \} ^{T}_{t=1}$ represents the original DDIM inversion chain, and $\left \{ \bar{z} _{t} \right \} ^{T}_{t=1}$ is the auxiliary inversion chain obtained by DCI.

During the inference process, we use $\bar{z} _{T}$ as the latent noise vector and $\left \{ \hat{z} _{t} \right \} ^{T}_{t=1}$ as the input for the noise prediction network at corresponding timesteps to compute the noise that needs to be removed, as shown in Eq.(\ref{eq:11}),
\begin{equation}
\Bar{z}_{t-1}=\sqrt{ \bar{\alpha}_{t-1} } \frac{\bar{z}_{t}-\sqrt{1-\bar{\alpha}_{t}} \tilde{\epsilon}(\hat{z}_{t},t)\ }{\sqrt{\bar{\alpha} _{t} } }+ \sqrt{1-\bar{\alpha}_{t-1}} \tilde{\epsilon}(\hat{z}_{t},t).
  \label{eq:11}
\end{equation}
By completing the inference process following Eq.(\ref{eq:11}), we can achieve a fully exact reversible diffusion process, thereby ensuring precise image reconstruction and fine image editing. 

The complete mathematical derivation about exact reversible diffusion is detailed in Alg.\ref{Alg:ERDDCI}. 
In the original DDIM framework, inversion is performed through Eq.(\ref{eq:8}) and inference is performed through Eq.(\ref{eq:9}), approximating the noise injected and removed from latent image variable at timestep $t$ to be the same based on $\tilde{\epsilon}(\hat{z}_{t-1},t) \approx \tilde{\epsilon}(z'_{t},t)$, thereby achieving an approximately reversible diffusion.

However, in our approach, we construct an auxiliary inversion chain $\left \{ \bar{z} _{t} \right \} ^{T}_{t=1}$, which is derived from $\left \{ \hat{z} _{t} \right \} ^{T}_{t=1}$ by Eq.(\ref{eq:10}). The inference process is then carried out by replacing Eq.(\ref{eq:9}) with Eq.(\ref{eq:11}), which removes the noise added at each timestep on the auxiliary inversion chain precisely (as shown in Fig. \ref{fig:overview}), so that the inference process reconstructs the original image exactly in the opposite direction of the auxiliary inversion chain without any offset, thus enabling a completely precise and reversible diffusion process.

The above operation realizes exact reversible diffusion: it replaces the native DDIM inference trajectory (DIT) corresponding to Eq.(\ref{eq:9}) with a joint inference trajectory (JIT) that uses both the DDIM inversion chain and the auxiliary inversion chain for inference. In the image reconstruction task (where the target prompt is the same as the original prompt), exact reversible diffusion means that the model can perfectly recover the inverted original image when $\omega=1$ and the number of diffusion steps is large enough (to minimize the model's rounding error). As shown in Tab.\ref{tab:recon_acc}, ERDDCI can achieve perfect reconstruction with LPIPS = 0.001 and SSIM = 0.999 for $\omega=1$ and timesteps = 50. And this kind of perfect reconstruction is not possible with approximately reversible diffusion like DDIM. In addition, when $\omega >1$, our method can also complete image reconstruction with significantly higher precision than other methods.

During image editing, guidance scale $\omega$ is usually not 1 and the target prompt needs to be rewritten according to the desired image effect.
Although ERDDCI achieves precise image reconstruction and provides support for the high fidelity of the edited images, since JIT will return to the original image to the maximum extent, this will impose some limitations on the flexibility of image editing, and we need some additional operations to ensure high-quality image editing.
We will elaborate on this in subsequent sections.

\begin{algorithm} [tb]
\caption{Exact Reversible Diffusion via DCI}
\textbf{Input}: $z_{0}$\\
\textbf{Symbol}: ``$\oplus$'' is the noise addition operation; ``$\ominus$'' is the noise removal operation; $\tilde{\epsilon}$ is the noise prediction networks; $\left \{ \hat{z} _{t} \right \} ^{T}_{t=1}$ is the original DDIM inversion chain; $\left \{ \bar{z} _{t} \right \} ^{T}_{t=1}$ is an auxiliary inversion chain assisting in achieving exact diffusion.\\
\textbf{Output}: $z^{*}_{0}$\\
\hspace*{\dimexpr\algorithmicindent*0\relax}\rule{1\linewidth}{0.4pt}\\ % Short horizontal line
\textbf{Stage 1: DCI Process}
\begin{algorithmic}[1]
\STATE Receive $z_{0}$
\STATE Initialize $\hat{z}_{0}\longleftarrow z_{0}$, $\Bar{z}_{0}\longleftarrow z_{0}$
\FOR{$t = 1$ to $T$}
    \STATE $\hat{z}_{t}\longleftarrow \hat{z}_{t-1} \oplus \tilde{\epsilon}(\hat{z}_{t-1},t)$
    \STATE $\bar{z}_{t}\longleftarrow \bar{z}_{t-1} \oplus \tilde{\epsilon}(\hat{z}_{t},t)$
\ENDFOR
\STATE \textbf{return} $\bar{z}_{T}$, $\left \{ \hat{z} _{t} \right \} ^{T}_{t=1}$
\end{algorithmic} 
\hspace*{\dimexpr\algorithmicindent*0\relax}\rule{1\linewidth}{0.2pt}\\ % Short horizontal line
\textbf{Stage 2: Inference Process}
\begin{algorithmic}[1]
\STATE Receive $\bar{z}_{T}$, $\left \{ \hat{z} _{t} \right \} ^{T}_{t=1}$
\FOR{$t = T$ down to $1$}
    \STATE $\bar{z}_{t-1} \longleftarrow \bar{z}_{t} \ominus \tilde{\epsilon}(\hat{z}_{t},t)$
\ENDFOR
\STATE \textbf{return} $z^{*}_{0} \longleftarrow \bar{z}_{0}$
\end{algorithmic}
\label{Alg:ERDDCI}
\end{algorithm}

\begin{table}[t!]
\centering
\begin{tabular}{@{}cccccc@{}}
\toprule
\multicolumn{1}{c}{Methods} & \multicolumn{1}{c}{PTP} & \multicolumn{1}{c}{NTI} & \multicolumn{1}{c}{AIDI} & \multicolumn{1}{c}{EDICT} & ERDDCI \\ \midrule
50 steps & 5.799 & 78.191 & 46.471 & 418.655 & 12.672 \\
20 steps & 2.469 & 33.768 & 18.803 & 170.120 & 4.986  \\
10 steps & 1.384 & 16.923 & 9.813  & 87.428  & 3.163  \\ \bottomrule
\end{tabular}
\caption{Averaged time consumed for reconstructing a single image by various inversion methods, measured in seconds.}
\label{tab:duration}
\end{table}

\subsection{Dynamic Control Strategy for Guidance Scale}
Although ERDDCI provides strong assurance for the fidelity of edited images, under high guidance scales ($\omega >1$), unavoidable rounding errors or overfitting of backbones, along with uncertainties introduced by target prompt, are magnified \cite{sadat2024eliminating}. To some extent, these factors can compromise the quality of image reconstruction and editing.
To optimize ERDDCI for ideal reconstruction and editing of real images at high guidance scales, we introduce a DCS for the guidance scales.
In PTP's editing approach, a static guidance scale is applied to image editing. 
This implies that during the initial phases of image generation, which are focused on establishing basic semantic structures, a higher guidance scale can readily disrupt the image's original semantic structure, aligning it more closely with the semantics described by the target prompt. 
This method often fails to balance our two criteria, resulting in undesired editing outcomes. To improve upon PTP's attention mechanism-based automatic masking method, we have made further refinements. For fine-grained image editing, we have implemented a dynamic approach to activating the guidance scale, allowing for a gradual and controlled adjustment, as shown below, 
%Often, this can not generate the intended effect.
%Building upon PTP's attention mechanism-based automatic masking method, we have further refined it. 
%For fine-grained image editing, we modify the timestep for activating the guidance scale in a dynamic manner,
\begin{equation}
Guidance=\left\{\begin{matrix}1,  & t\ge  \sigma \\ L(Guidance,t), & otherwise\end{matrix}\right.,
  \label{eq:12}
\end{equation}
where $\sigma$ represents the timestep at which the guidance scale is activated, controlled by the control factor $\eta$, where $\sigma = (1-\eta)T, \eta \in [0,1]$.
And then gradually linearly increasing the scale after activation until the pre-set level is reached: 
\begin{equation}
L(Guidance,t)=1+\frac{Guidance-1}{\sigma-t_{end}} (\sigma-t),
  \label{eq:13}
\end{equation}
$t_{end}$ controls the timestep at which the pre-set guidance scale is reached.
This editing process is more congruent with the mechanism of the image generation process \cite{biroli2024dynamical}. 
It allows for the modification or addition of new semantic information that ``matches in intensity'' with the semantics of the image already generated at the current timestep as the image's semantic content progressively fills in. 
% In this way, DCS avoids significant semantic drift in the edited image relative to the original image, ensures fidelity between the edited image and the original image while modifying the semantic information of the image.
In this way, DCS prevents major semantic drift, ensuring the reconstructed or edited image remains similar to the original while adding or modifying its semantic content.

By integrating DCI with DCS, we can artificially minimize the error accumulation caused by high guidance scales  and make the inference path as close as possible to the inference path of exact reversible diffusion (i.e., the inference path when $\omega = 1$). 
This ensures that real images can be precisely reconstructed and finely edited under high guidance scales.

\subsection{Real Image Editing by ERDDCI}
The difference between image editing and image reconstruction is that after inverting the original image, in the inference process, image reconstruction only needs to input the same target prompts as the original prompts, while image editing needs to rewrite the target prompts according to the requirements to achieve the desired editing effect. We focus on three types of edits: object transformation (T1), object refinement (T2), and style transfer (T3), with examples of target prompt rewrites for these editing types shown in the Tab. \ref{tab:prompt}. By inputting the rewritten target prompt and the inverted noise variables into the model for inference, the edited image can be produced.

\begin{table}[h]
\centering
\begin{tabular}{@{}c p{3cm} p{3cm}@{}}
\toprule 
\begin{tabular}[c]{@{}c@{}}Editing\\ types\end{tabular} & \multicolumn{1}{c}{Source prompt} & \multicolumn{1}{c}{Target prompt} \\
\midrule 
T1            & \textit{A cat sitting next to a mirror.} & \textit{A \textbf{dog} sitting next to a mirror.}                       \\
T2            & \textit{A cat sitting next to a mirror.} & \textit{A \textbf{sculptural} cat sitting next to a mirror.}            \\
T3            & \textit{A cat sitting next to a mirror.} & \textit{\textbf{Watercolor drawing of} a cat sitting} \textit{next to a mirror.} \\
\bottomrule 
\end{tabular}
\caption{Example of a target prompt rewrites.}
\label{tab:prompt}
\end{table}

Since pretrained models are trained based on the DDIM, the latent image variables on the native DIT are better adapted to the model's processing of variables, meaning they can generate semantic information that better matches the input prompt. However, since DDIM adheres to a local linearity assumption, errors will accumulate progressively during inference. If the guidance scale is high, these errors increase further, leading to semantic drift in the edited images——although they match the target prompt’s description, they can deviate significantly from the original image's semantic content, rendering the edits meaningless.

On the other hand, if the auxiliary chain $\left \{ \bar{z} _{t} \right \} ^{T}_{t=1}$ and DDIM inversion chain $\left \{ \hat{z} _{t} \right \} ^{T}_{t=1}$ are used for joint inference as described in Alg.1, the latent image variables along this JIT will be strongly constrained by the auxiliary chain, striving to maintain high similarity between the edited and original images. Additionally, since the noise predicted at each timestep on the auxiliary chain is not based on the latent image variables from the previous timestep ($\bar{z}_{t+1}$) but rather on those from the corresponding timestep of the DDIM inversion chain ($\hat{z}_{t}$), when the guidance scale is high, images generated along this JIT are highly likely to exhibit artifacts.

To harness the strengths and counterbalance the limitations of both DIT and JIT, we redesigned the image editing inference process with DCS, as shown in Alg. \ref{Alg:ERDDCIedit}. Here, $JI$ stands for joint inference:
\begin{algorithm}[tb]
\caption{Image Editing Inference Process}
\textbf{Input}: $\bar{z}_{T}$, $\left \{ \hat{z} _{t} \right \} ^{T}_{t=1}$\\
\textbf{Symbol}: $JI$ represents joint inference, which conformed to Alg.1; $DJI$ represents combining DDIM inference  mechanisms with $JI$; $\sigma$ is activated timestep; $m+n=1$ is the dependency factors; $\Omega$ is guidance scale from users; $\omega$ is guidance scale actually input into model; $r$ controls the number of timesteps $DJI$ is executed with greater reliance on the auxiliary chain; $z^{\star}_{0}$ represents the final edited image. \\
\textbf{Output}: $z^{\star}_{0}$
\begin{algorithmic}[1]
\STATE \textbf{Receive} $\bar{z}_{T}$, $\left \{ \hat{z} _{t} \right \} ^{T}_{t=1}$
\STATE Initialize $\breve{z}_{T}\longleftarrow \bar{z}_{T} $
\FOR{$t = T$ down to $1$}
    \IF{$t \ge  \sigma$}
        \STATE $\omega=1$
        \STATE $\breve{z} _{t-1} = JI(\omega,\breve{z}_{t},\hat{z} _{t} )$
    \ELSE
        \IF{$i \bmod r = 1$}
            \STATE $\omega=1, ~m=0.8, ~n=0.2$
            \STATE $\breve{z} _{t-1} = DJI(\omega,\breve{z} _{t},\hat{z} _{t}, m, n )$
        \ELSE
            \STATE $\omega=L(\Omega,t), ~m=0.5, ~n=0.5$
            \STATE $\breve{z} _{t-1} = DJI(\omega,\breve{z} _{t},\hat{z} _{t}, m, n )$
        \ENDIF  
    \ENDIF
\ENDFOR
\STATE \textbf{return} $z^{\star}_{0} \longleftarrow\breve{z}_{0} $

\end{algorithmic}
\label{Alg:ERDDCIedit}
\end{algorithm}

\begin{equation}
JI(gs,\breve{z} _{t},\hat{z} _{t} )= Att( g(\breve{z}_{t},t,\Tilde{\epsilon}(\hat{z} _{t},t,\omega ) ) ),
  \label{eq:14}
\end{equation}
where $\breve{z}_{t}$ stands for latent image variable in image editing inference process and set $\breve{z}_{T}= \bar{z}_{T}$, $Att(\cdot )$ represents the modification operation on attention weights, $g(\breve{z}_{t},t,\Tilde{\epsilon})$ denotes the latent image variable at timestep $t-1$ , with its attention weights yet to be modified:
\begin{equation}
g(\breve{z}_{t},t,\Tilde{\epsilon})=\sqrt{ \bar{\alpha}_{t-1} } \frac{\breve{z}_{t}-\sqrt{1-\bar{\alpha}_{t}} \tilde{\epsilon}\ }{\sqrt{\bar{\alpha} _{t} } }+ \sqrt{1-\bar{\alpha}_{t-1}} \tilde{\epsilon},
  \label{eq:15}
\end{equation}
and $\Tilde{\epsilon}$ conformed to Eq.(\ref{eq:5}). $DJI$ represents the inference that combines JIT with DIT:
\begin{equation}
 DJI(\omega,\breve{z} _{t},\hat{z} _{t}, m, n )= Att( g(\breve{z}_{t},t,\Tilde{\epsilon}((m\hat{z} _{t}+n\breve{z} _{t}),t,\omega ) ) ).
  \label{eq:16}
\end{equation}
$DJI$ makes the noise predicted at each step is determined by both the latent image variables from the previous timestep in the inference trajectory and those from the corresponding timestep on the DDIM inversion chain. The parameters $m$ and $n$ are used to adjust the predicted noise’s reliance on these two variables. To enhance efficiency, during most timesteps of the inference process (Alg.2, line 10), we set $m$ and $n$ to 0.5, indicating equal dependency. In a few timesteps (Alg.2, line 8), we adjust $m$ to 0.8 and $n$ to 0.2, emphasizing a greater reliance on the constraints from the auxiliary chain. Users can also finely adjust $m$ and $n$ for different images to generate higher quality images. 

\begin{figure}[t!]
  \centering
  \includegraphics[width=\columnwidth]{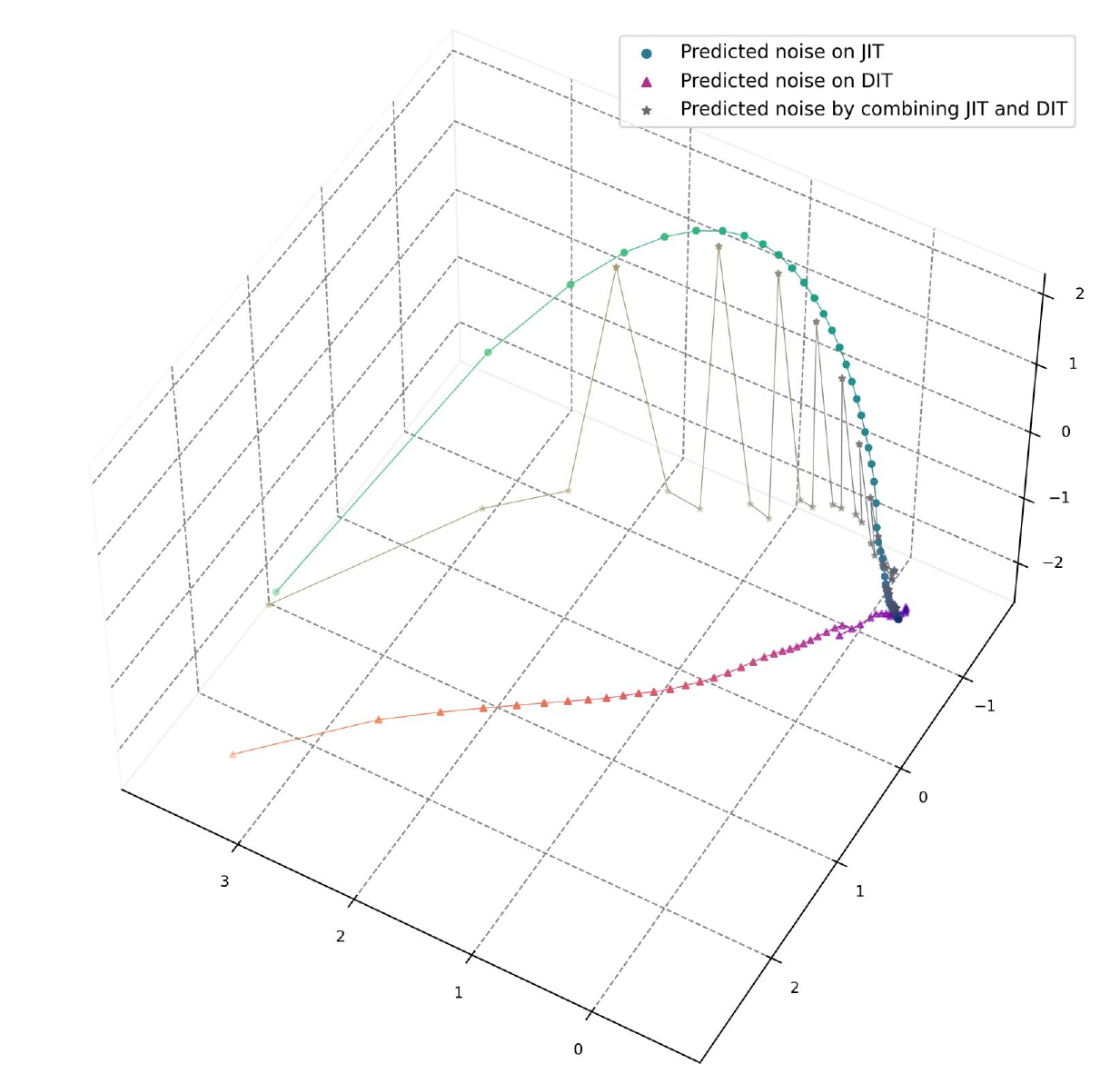}
  \caption{{\bf Predicted Noise on Different Inference Trajectories over Timesteps.} The color of the trajectory gradually becomes lighter with the reverse timesteps ($T \longrightarrow 0$).
  }
  \label{fig:noisevil}
\end{figure}

To ensure the layout structure of the edited image remains consistent with the original, we set the activated timestep $\sigma$. During the early inference stages (when the timestep $t \ge  \sigma$),  $\sigma$ is consistently set to 1, and joint inference is used to ensure precise semantic alignment with the original image. In the later stages of the inference process (when the timestep $t < \sigma$), we begin to incorporate the semantic information required by the target prompt, gradually aligning the image with the descriptions of the target prompt.  To further constrain potential semantic drift in the later stages of the inference process, certain timesteps performing inference that relies more on auxiliary chain, ensuring high fidelity between the edited and original images. 
As shown in Fig.\ref{fig:noisevil}, the noise predicted by $DJI$ falls between JIT and DIT to maintain good edit ability and fidelity to the original image. In the later stages of inference, more reliance on JIT is set at certain timesteps to maximally avoid unnecessary semantic drift in the edited images. Through such an inference process, ERDDCI can achieve ideal image editing results.

\section{Experiment}
All experiments in this paper are based on pre-trained Stable Diffusion, utilizing NVIDIA GeForce RTX 4090 GPU cards. 
We use 5000 test images from the COCO dataset \cite{lin2014microsoft} for real image reconstruction testing. 
For image editing, all real image examples are sourced from Behance, a global platform for visual creativity and sharing, rather than standard datasets. 
These images provide an in-depth evaluation of our method's generalizability and capability in handling complex real-world images.
We compare our method with other improved real image inversion and editing methods based on DDIM inversion, including PTP \cite{hertz2022prompt}, NTI \cite{mokady2023null}, EDICT \cite{wallace2023edict}, and AIDI \cite{pan2023effective}, to evaluate reconstruction and editing quality. Additionally, we verify the Generalizability of ERDDCI by replacing backbone with LDM. Finally, we visualize DIT and JIT to further analyze the effectiveness of our approach.
\subsection{Real Image Reconstruction}
\begin{figure}[t!]
  \centering
  \includegraphics[width=\columnwidth, height=0.5\columnwidth]{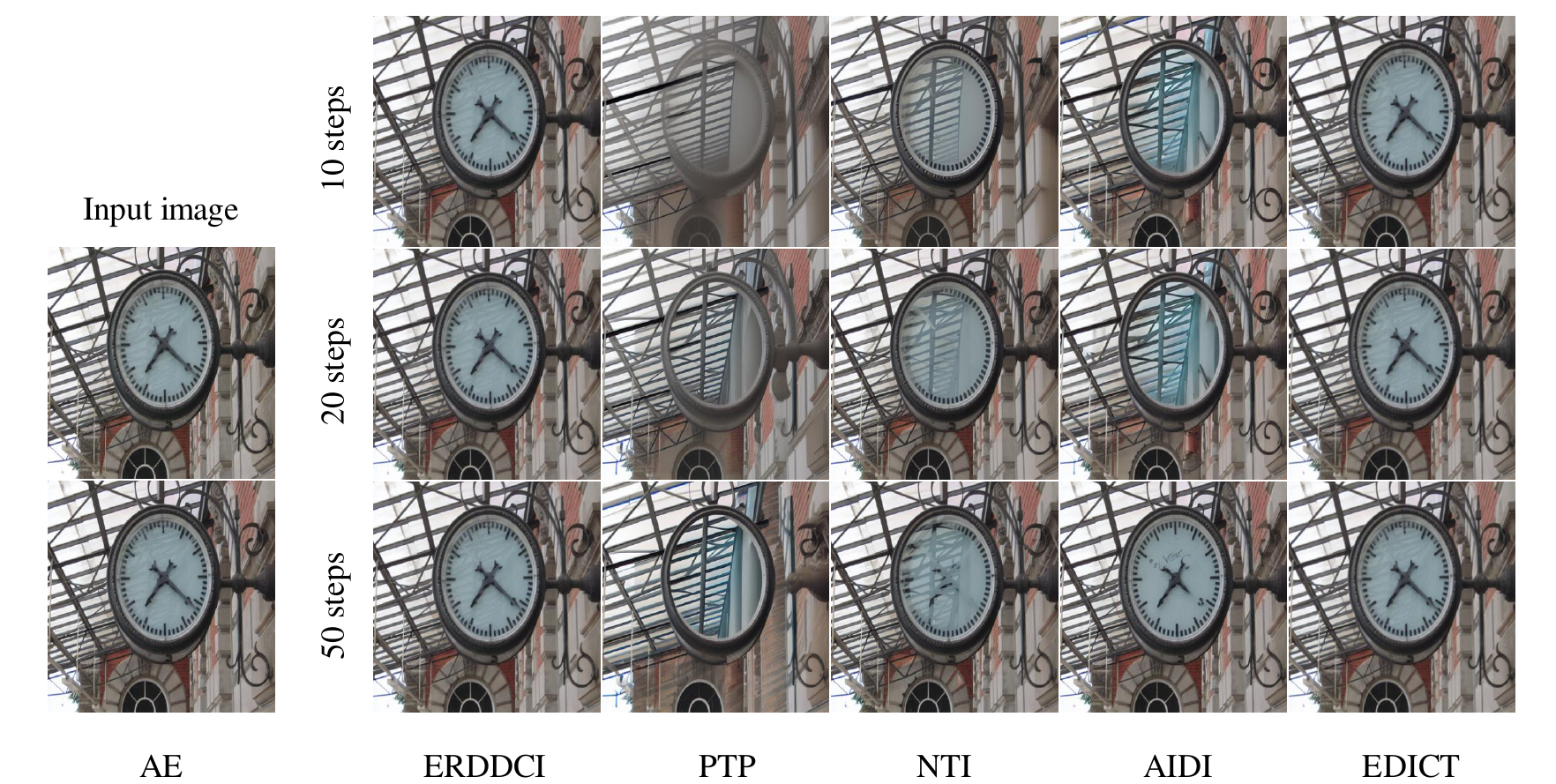}
  \caption{{\bf Reconstruction effects for various diffusion inversion methods.} The AE image is decoded from Stable Diffusion without inversion and used as the ground truth for other methods' reconstruction. 
  }
  \label{fig:reconstruction}
\end{figure}
\begin{figure}[t!]
  \centering
  \includegraphics[width=\columnwidth, height=0.6\columnwidth]{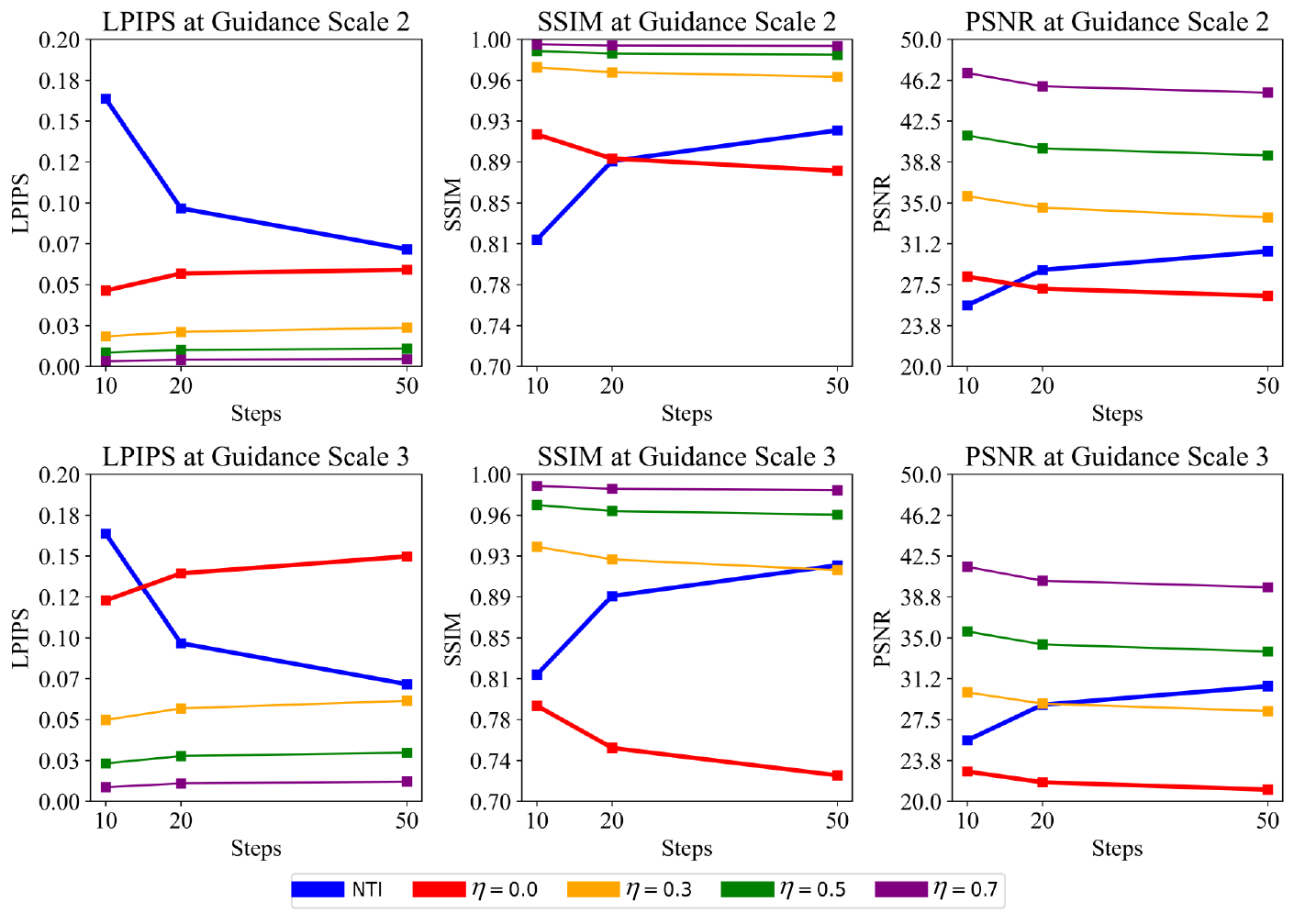}
  \caption{{\bf Quantitative evaluation.} %Quantitative evaluation of image reconstruction results for ERDDCI with DCS, NTI which optimizes the reconstruction under high guidance scales.
  Quantitative evaluation of image reconstruction results for ERDDCI with DCS and NTI. These two methods optimize the reconstruction under high guidance scales.
  }
  \label{fig:dynamiccompare}
\end{figure}

We conducted reconstruction tests using 5,000 images from the COCO2017 validation set, employing the first caption as the textual prompt. Quantitative results are illustrated in Tab.\ref{tab:recon_acc}. 
Compared with other optimized inversion methods under the DDIM framework, our proposed ERDDCI significantly improves the precision across various classifier-free guidance scales and diffusion steps.
For AIDI, we attempted to replicate the code based on its principles but could not verify its exact match with the original paper. 
Therefore, the quantitative data we provided about AIDI, as well as the visual examples in  Fig.\ref{fig:reconstruction}, are for reference only.
Other methods are tested and compared under the same conditions based on their open-source code.
\begin{table*}[t]
\centering
 \resizebox{\textwidth}{!}{
\begin{tabular}{@{}ccccccccccccccccc@{}}
\toprule
\multicolumn{2}{c}{} & \multicolumn{3}{c}{PTP} & \multicolumn{3}{c}{NTI} & \multicolumn{3}{c}{AIDI} & \multicolumn{3}{c}{EDICT} & \multicolumn{3}{c}{ERDDCI}  \\
\cmidrule{3-17}
\multicolumn{2}{c}{steps/scales} & ~10~ & ~20~ & ~50~ & ~10~ & ~20~ & ~50~ & ~10~ & ~20~ & ~50~ & ~10~ & ~20~ & ~50~ & ~10~ & ~20~ & ~50~ \\
\midrule
\multirow{3}{*}{\parbox[c]{1cm}{\centering\rotatebox[origin=c]{90}{LPIPS}}} & 1 & 0.222 & 0.175 & 0.135 & - & - & - & \underline{0.121} & \underline{0.117} & \underline{0.118} & 0.153 & 0.153 & 0.153 & {\bf0.003} & {\bf0.003} & {\bf0.001} \\
& 2 & 0.253 & 0.251 & 0.285 & 0.164 & \underline{0.097} & \underline{0.072} & 0.157 & 0.166 & 0.154 & \underline{0.153} & 0.153 & 0.153 & {\bf0.046} & {\bf0.056} & {\bf0.059} \\
& 3 & 0.304 & 0.331 & 0.373 & \underline{0.151} & {\bf0.083} & {\bf0.044} & 0.238 & 0.287 & 0.342 & 0.153 & 0.153 & 0.153 & {\bf0.123} &\underline{0.139} & \underline{0.150} \\

\multirow{3}{*}{\parbox[c]{1cm}{\centering\rotatebox[origin=c]{90}{SSIM}}} & 1 & 0.763 & \underline{0.810} & \underline{0.858} & - & - & -  & \underline{0.773} & 0.791 & 0.817 & 0.699 & 0.699 & 0.699 & {\bf0.997} & {\bf0.997} & {\bf0.999} \\
& 2 & 0.723 & 0.721 & 0.692 & \underline{0.816} & \underline{0.888} & {\bf0.917} & 0.716 & 0.714 & 0.705 & 0.699 & 0.699 & 0.699 & {\bf0.913} & {\bf0.891} & \underline{0.879} \\
& 3 & 0.662 & 0.629 & 0.587 & {\bf0.823} & {\bf0.899} & {\bf0.944} & 0.661 & 0.637 & 0.659 & 0.699 & 0.699 & 0.699 & \underline{0.788} & \underline{0.749} & \underline{0.724} \\

\multirow{3}{*}{\parbox[c]{1cm}{\centering\rotatebox[origin=c]{90}{PSNR}}} & 1 & 23.852 & 24.848 & \underline{26.710} & - & - & - & 24.694 & 24.867 & 25.427 & \underline{25.784} & \underline{25.784} & 25.784 & {\bf65.578} & {\bf65.822} & {\bf66.641} \\
& 2 & 22.230 & 21.566 & 20.352 & 25.587 & {\bf28.852} & {\bf30.562} & 21.311 & 21.175 & 21.458 & \underline{25.784} & 25.784 & 25.784 & {\bf28.235} & \underline{27.124} & \underline{26.457} \\
& 3 & 20.293 & 18.987 & 17.789 &\underline{25.646} & {\bf29.513} & {\bf33.226} & 20.287 & 19.251 & 19.463 & {\bf25.784} & \underline{25.784} & \underline{25.784} & 22.735 & 21.743 & 21.070 \\
\bottomrule
\end{tabular}
}
\caption{Testing the precision of reconstruction across various inversion methods, this analysis includes outcomes from diverse inversion steps and multiple classifier-free guidance scale settings. The evaluation employs three key perceptual metrics: LPIPS, SSIM and PSNR, Lower values are more desirable for LPIPS, while higher values are better for SSIM and PSNR.}
\label{tab:recon_acc}
\end{table*}

\begin{figure*}[t!]
  \centering
  \includegraphics[width=\textwidth]{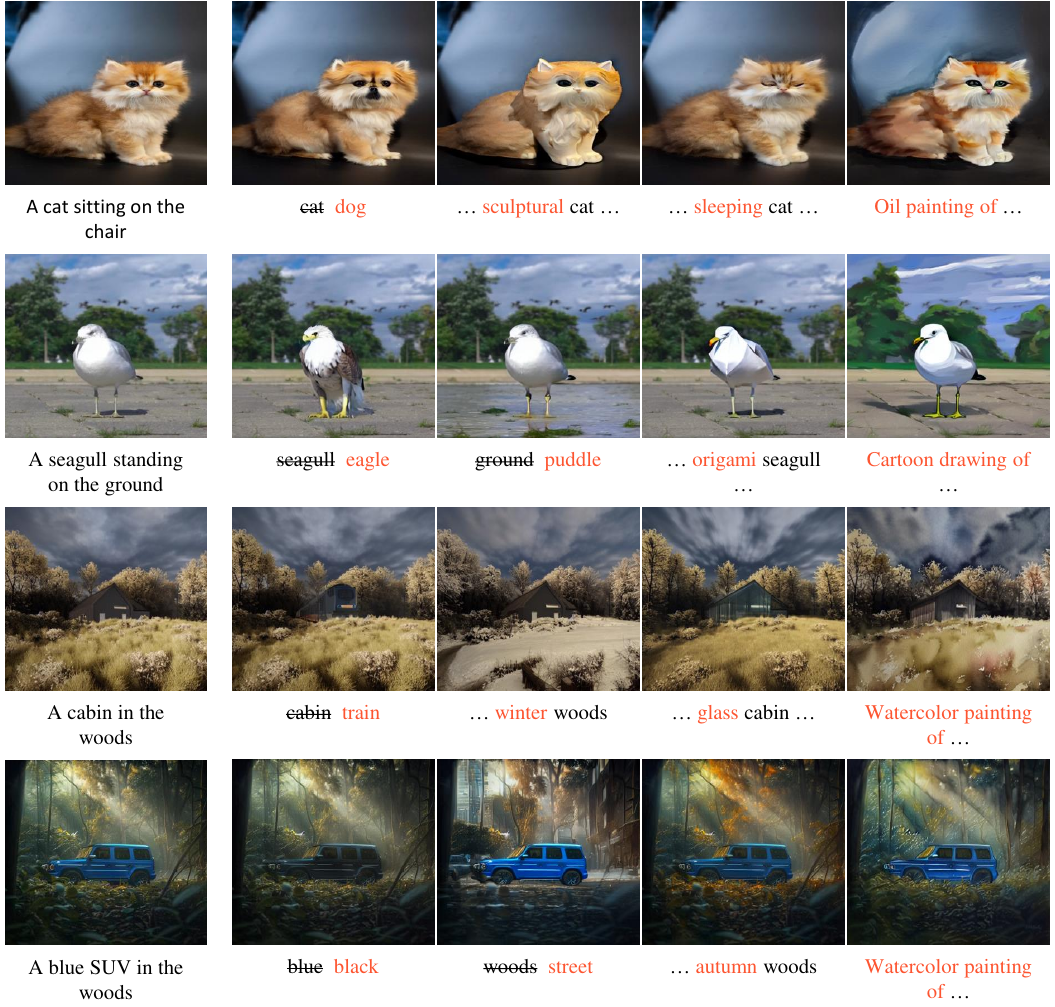}
  \caption{{\bf Visual examples of real image editing.} The left side is the real image to be edited, and the four columns on the right are the edited images. Users define all prompts. ERDDCI performs 20 editing steps with DCS. 
  }
  \label{fig:imageedit}
\end{figure*}

Fig.\ref{fig:reconstruction} showcases a complex real image featuring intricately interwoven lines in the background, a detailed clock face, and a reflective glass surface of the clock. 
These elements represent challenges that existing inversion methods struggle to address effectively. 
As observed in Fig.\ref{fig:reconstruction}, ERDDCI successfully reconstructs the image in just 10 steps. According to the quantitative analysis in Tab.\ref{tab:recon_acc}, ERDDCI can reach LPIPS = 0.001 and SSIM = 0.999 under $\omega=1$, indicating that it is almost identical to the ground
truth.
This demonstrates that ERDDCI achieves a truly exact image inversion and reconstruction.
The reconstruction results of EDICT are visually similar to those of our approach but exhibit inferior quantitative outcomes and a notably longer image processing time.
AIDI produces acceptable reconstruction results at 50 steps, but visible artifacts can be seen on the clock face in the image. 
In contrast, PTP and NTI are nearly ineffective in reconstructing such complex images.

\begin{figure*}[t!]
  \centering
  \includegraphics[width=\textwidth]{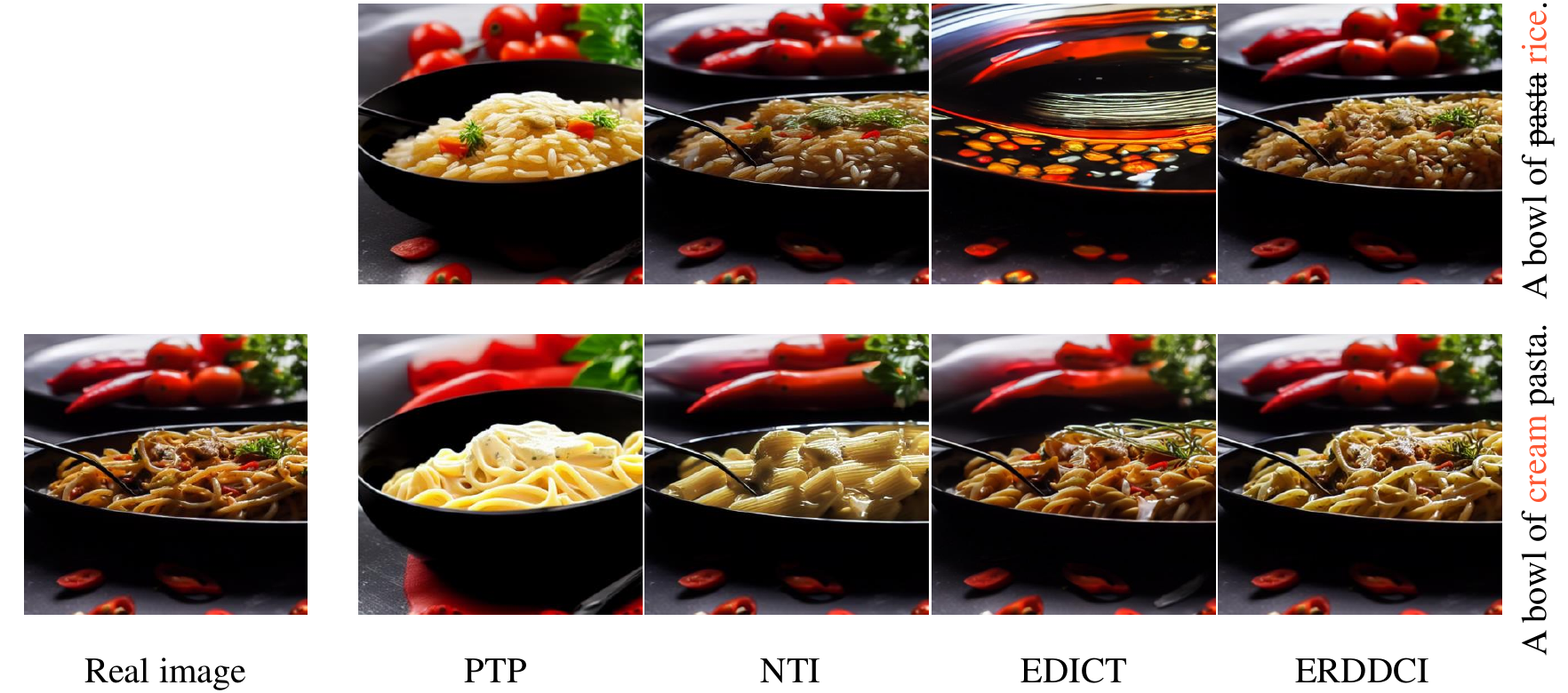}
  \caption{{\bf Image editing effects between ERDDCI and other methods.} ERDDCI can achieve an excellent balance between preserving the semantic information of the original image and conforming to the semantics of the target prompt.
  }
  \label{fig:editcompared}
\end{figure*}
To further enhance the reconstructed precision of images under high guidance scales, we incorporate DCS into the image reconstruction process, which significantly improved reconstructed outcomes. 
The effectiveness of DCS is demonstrated in Fig.\ref{fig:dynamiccompare}, where we observe a clear trend: as the control factor $\eta$ increases, the evaluation metrics of the images reconstructed by ERDDCI surpass those obtained by NTI, which has been optimized for high guidance scales.  
This trend suggests that DCS can ensure high-quality image reconstruction under high guidance scales while maintaining a significant advantage in computational speed and efficiency, as seen in Tab.\ref{tab:duration}.
DCS enhances the robustness of the inference process against variations in guidance scales. 
By dynamically adjusting the guidance scale based on the specific requirements of the image content and the desired level of detail, our method can cater to a broader range of image reconstruction tasks, ensuring high-quality outcomes across diverse scenarios.

\subsection{Real Image Editing}
Partial examples of real image editing using ERDDCI %as demonstrated 
are shown in Fig.\ref{fig:imageedit}. 
We focus on three image editing tasks: object transformation, object fine-tuning, and style transfer with global image. 

\begin{figure}[t!]
  \centering
  \includegraphics[width=\columnwidth]{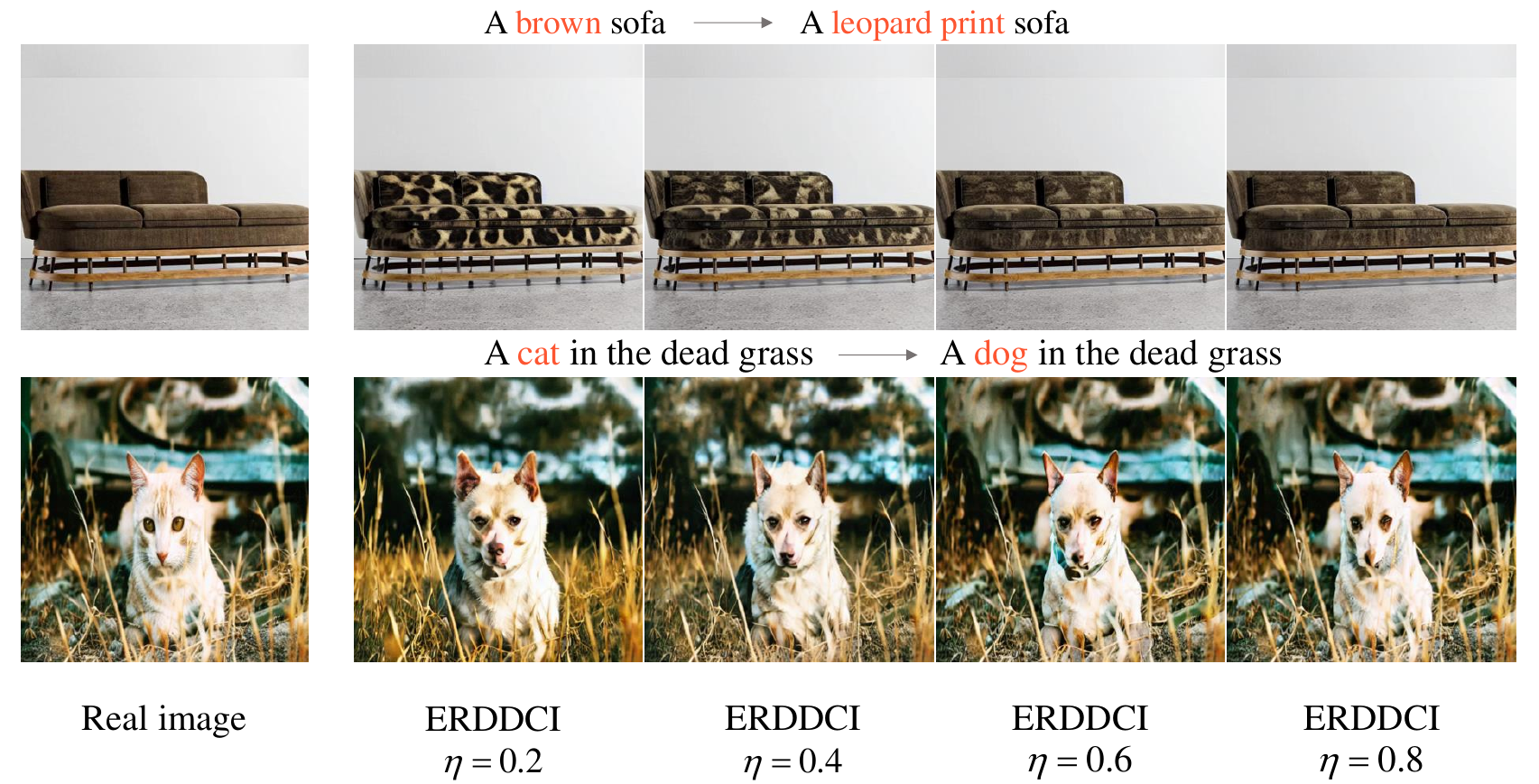}
  \caption{{\bf Image editing effects under DCS.} Edit the cat into a dog, where $\eta$ is the control factor, ranging from [0, 1]. The guidance scale is uniformly set to 7.5, and $t_{end}=T$.
  }
  \label{fig:dynamic}
\end{figure}

Across all examples, ERDDCI can consistently preserve the semantics of unedited image parts, leaving them unaffected. 
For edited parts, ERDDCI can maintain appropriate semantic consistency with the original, as seen in the first row where a cat transformed into a dog retains similar fur texture and color, and when fine-tuning into a sculpture, maintains the original's contour and pose. 
However, in the third row, ERDDCI makes the edited train share almost no semantic similarity with the original cabin.
This demonstrates ERDDCI's versatility, which can either preserve or entirely discard the original image semantics as needed. 
The underlying flexibility is due to our DCS, which allows for adjustable semantic consistency tailored to specific editing. 
By adjusting the control factor $\eta$, the semantic similarity between edited objects and the original can be controlled, maintaining an appropriate connection between the edited and original images.
As illustrated in Fig.\ref{fig:dynamic}, our method's editing outcomes under varying control factors reveal that higher values of $\eta$ result in greater semantic similarity to the original. 
Take the picture below as an example. The dog's facial contours becoming increasingly similar to a cat, the ears gradually developing white fur similar to that in the cat's ears from the original image, and in extreme cases (\textit{e.g.},$\eta=0.8$), the dog even grows whiskers like the cat in the original image.

\subsection{Comparison with the Baseline}
Fig.\ref{fig:editcompared} compares the editing effects of our method with those of other mainstream baseline approaches. 
ERDDCI strikes a perfect balance between preserving the original image's semantics and aligning with the target prompt's semantics, and exhibits greater realism and superior detail handling.  
For instance, in object transformation editing tasks, images edited with ERDDCI exhibit a more realistic and clear texture of rice grains compared to NTI, while maintaining other semantic information like the beef on the pasta unchanged. 
In the task of object refinement editing, ERDDCI meticulously adjusts the color and texture of pasta, imparting a creamy texture. 
Although NTI also edits according to the target prompt, it neglects the original image's semantic information and changes the type of pasta, which generally deviates from user expectations.
A similar issue occurs with PTP, where edited images tend to align more closely with the target prompt's semantics. 
As for EDICT, it demonstrates instability in editing tasks and fails to accurately interpret the target prompt, for instance, not incorporating any creamy elements into the image and failing to maintain semantic consistency in unedited sections. 

\begin{figure}[t!]
  \centering
  \includegraphics[width=\columnwidth]{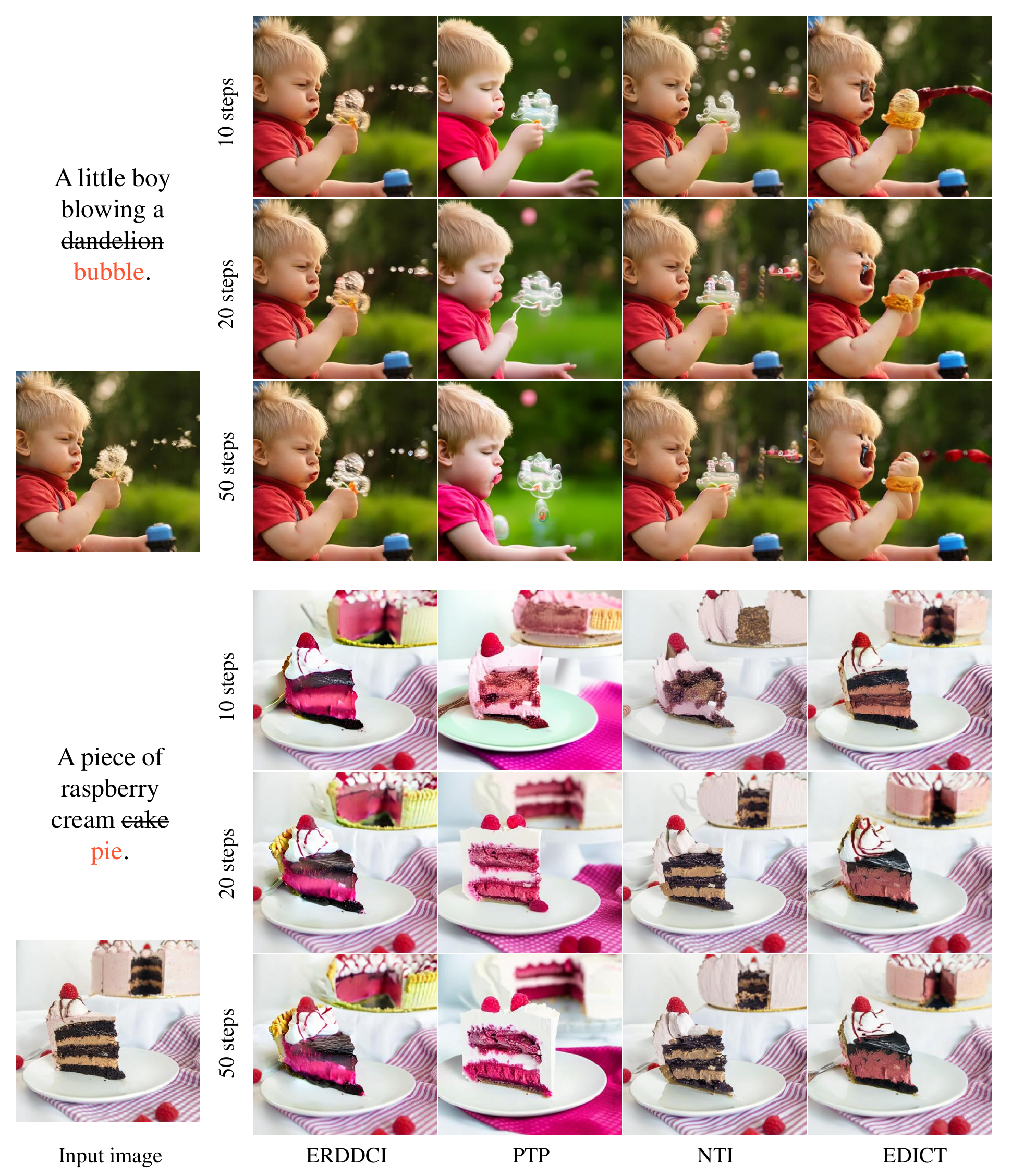}
  \caption{{\bf Image editing effects between ERDDCI and other methods at different timesteps}The left side is the real image to be edited, and the four columns on the right are the edited images from different methods, each image shows the editing effect of 10, 20, and 50 timesteps respectively. All prompts are defined by users.  }
  \label{fig:edit_step}
\end{figure}

Additionally, We qualitatively compared the editing effects of the various methods at different time steps. As shown in Fig. \ref{fig:edit_step}, these comparisons reveal that ERDDCI delivers the desired editing effects at almost all the timesteps, showcasing its stability and reliability in image editing tasks. In contrast, the PTP method fails to achieve high fidelity, while the editing outcomes of NTI and EDICT demonstrate unstable.

In addition to qualitative comparisons evaluating the image editing effectiveness of the different methods, we also performed detailed quantitative comparisons to provide a more rigorous and objective measure of performance. Given the absence of a universally recognized dataset for quantitative analysis of generic images and editing tasks, we adopted the established methods from the AIDI and EDICT projects, ensuring a fair and consistent basis for comparison. Specifically, we selected five animal classes from the ImageNet dataset, with class IDs 386, 348, 285, 294, and 185, corresponding to an elephant, a male sheep, a cat, a brown bear, and a dog, respectively. We performed three types of edits on these images: changing the subject, altering the background, and style transfer. To objectively measure the fidelity of the edited images to the originals, we utilized the LPIPS, which quantifies perceptual similarity. Additionally, we employed the CLIP to assess how well the edits align with the target prompt. The results, as shown in Tab.\ref{tab:qa}, indicate that our method achieved the best scores at nearly all timesteps.

\begin{table}[t]
\centering
 \resizebox{\columnwidth}{!}{
\begin{tabular}{@{}ccccccc@{}}
\toprule
\multicolumn{2}{c}{methods/timesteps} & \multicolumn{1}{c}{PTP} & \multicolumn{1}{c}{NTI} & \multicolumn{1}{c}{AIDI} & \multicolumn{1}{c}{EDICT} & \multicolumn{1}{c}{ERDDCI}  \\

\midrule
\multirow{3}{*}{\parbox[c]{1cm}{{~~~~~~LPIPS}}} & 10 & 0.421 & 0.459 & 0.258 & {\bf0.221} & 0.231  \\
& 20 & 0.445 & 0.461 & 0.315 & 0.291 & {\bf0.214}  \\
& 50 & 0.507 & 0.503 & 0.345 & 0.290 & {\bf0.206}\\

\multirow{3}{*}{\parbox[c]{1cm}{{~~~~~~~CLIP}}} & 10 & 0.475 & 0.296 & 0.682 & 0.355 & {\bf0.731} \\
& 20 & 0.522 & 0.372 & 0.733 & 0.578 & {\bf0.787}  \\
& 50 & 0.485 & 0.407 & 0.753 & 0.675 & {\bf0.789} \\

\bottomrule
\end{tabular}
}
\caption{Quantitative assessments for image editing using LPIPS and CLIP, Lower values are more desirable for LPIPS, while higher values are better for CLIP.}
\label{tab:qa}
\end{table}

\begin{figure*}[tbp]
  \centering
  \includegraphics[width=\textwidth, height=0.57\textwidth]{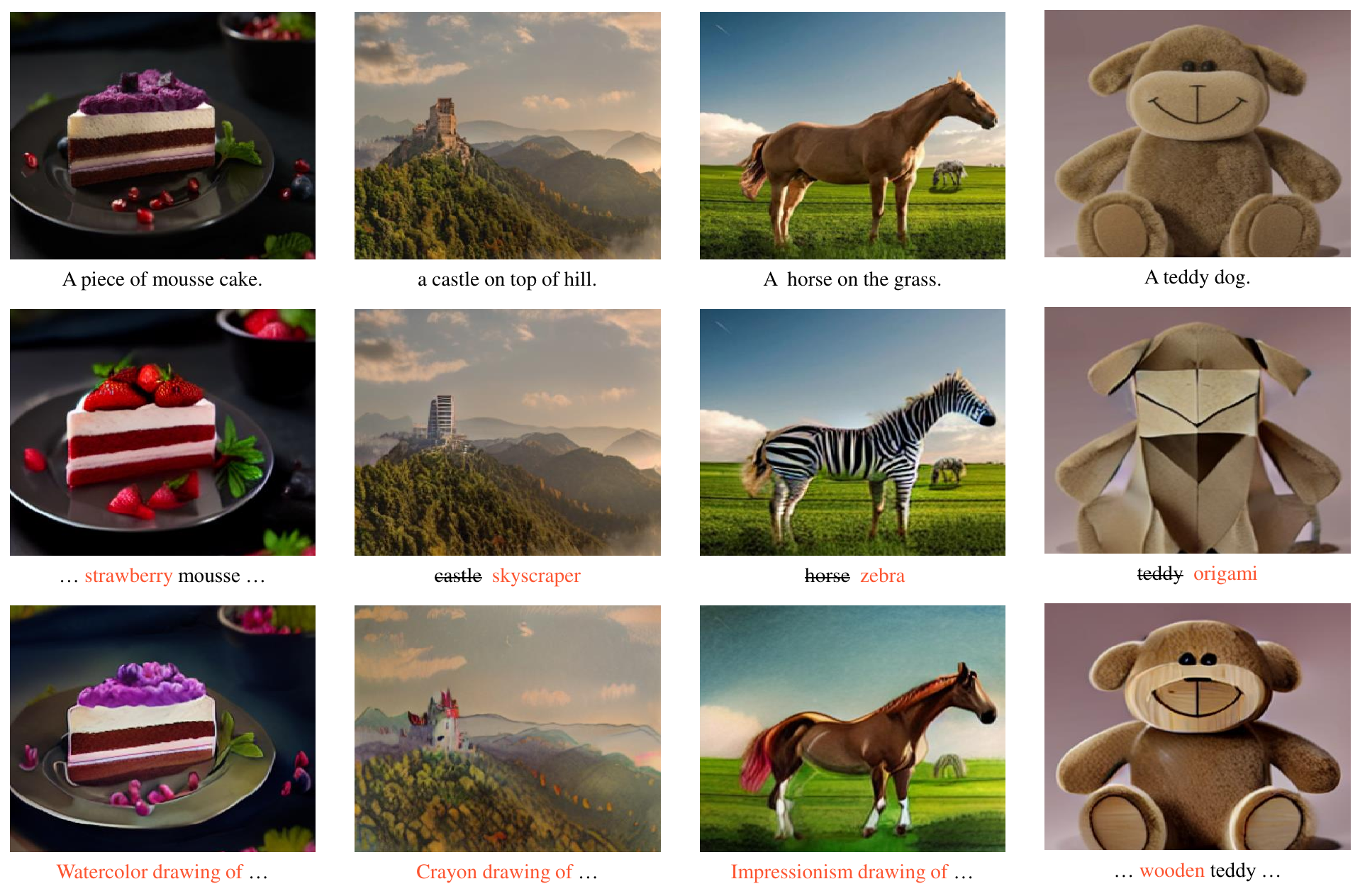}
  \caption{{\bf Visual examples of real image editing on LDM.} The first raw is the real image to be edited, and the remaining rows are the edited images. All prompts are defined by users. ERDDCI performs 50 editing steps with DCS.
  }
  \label{fig:ldm}
\end{figure*}
\subsection{Generalizability of ERDDCI}
ERDDCI, designed as a plug-and-play technique for inversion and editing, can seamlessly integrates with any model that employs diffusion mechanisms to manipulate images. To demonstrate the generalizability and robustness of ERDDCI across various architectures, we replaced the original backbone with the Latent Diffusion Model (LDM). This switch serves to further explore ERDDCI's capabilities in image editing under different operational conditions.This strategic substitution not only showcases ERDDCI’s adaptability across different diffusion-based architectures but also underscores its significant potential for widespread application in diverse image editing tasks. By employing LDM as the new backbone, we highlight ERDDCI's flexibility in handling different diffusion models, ensuring that the integrity and quality of image manipulation are maintained even when the underlying technology changes. Fig.\ref{fig:ldm} provides a visual example of image editing using LDM as the backbone, showcasing the effectiveness and general applicability of the ERDDCI approach in adapting to various underlying diffusion models without compromising the quality of the image manipulation.

\subsection{Cumulative Error and Semantic Constraints}
\begin{figure}[t!]
  \centering
  \includegraphics[width=\columnwidth]{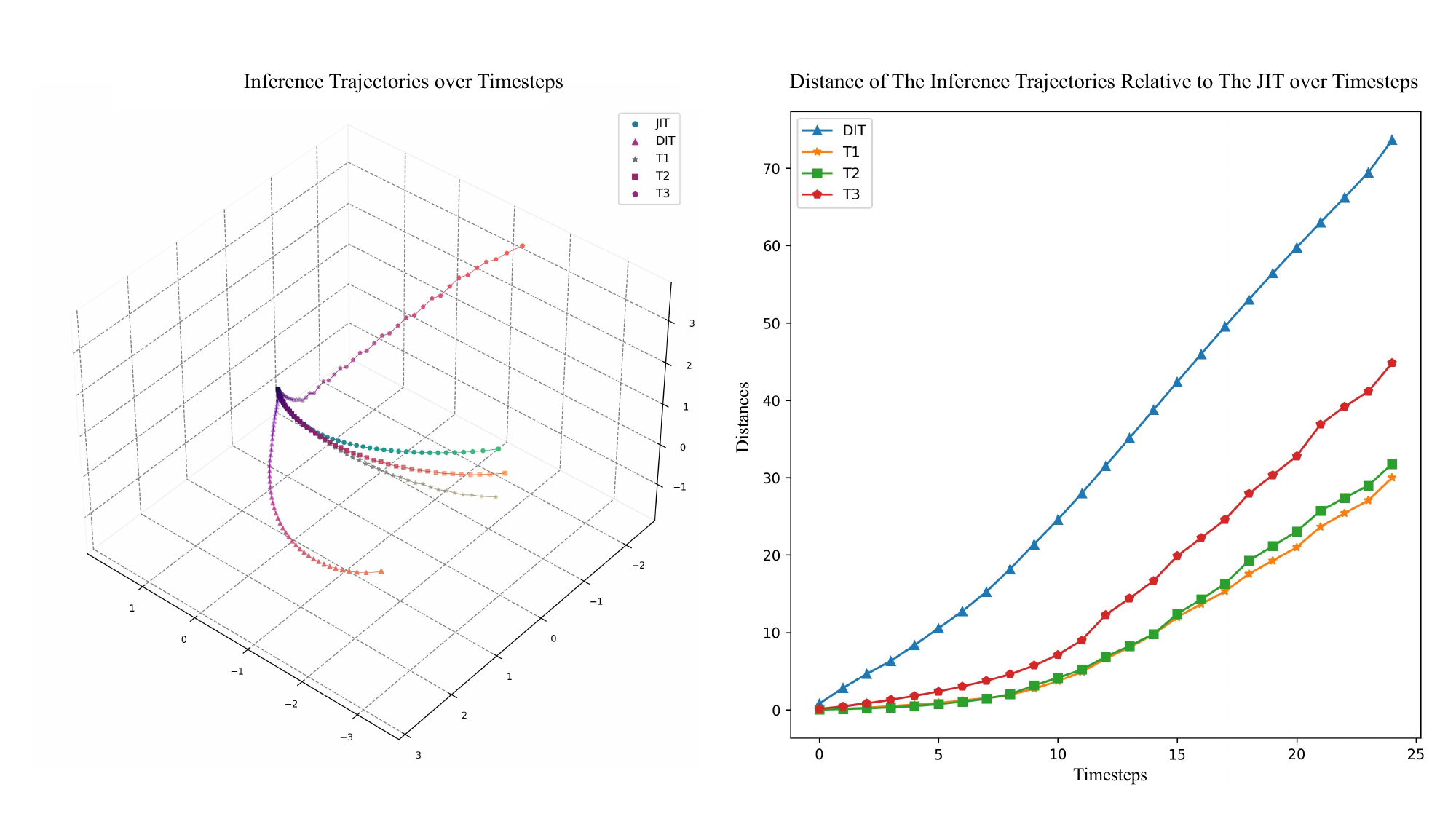}
  \caption{{\bf Visualization of Inference Trajectories and Semantic Drift .}  The left shows the reconstructed JIT, DIT, and the inference trajectory of the three edit forms. The color of the trajectory gradually becomes lighter with the reverse timesteps ($T \longrightarrow 0$). The right visualizes the semantic drift of other inference trajectories relative to JIT over timesteps.
  }
  \label{fig:errorvil}
\end{figure}
Joint inference using the auxiliary chain produced by DCI and the DDIM inversion chain allows for precise reconstruction of the original image. We can consider this JIT as an unbiased reference baseline. As shown in Fig.\ref{fig:errorvil}, we perform PCA dimension reduction on the latent image variables along the inference trajectory and then visualize them. Compared to DIT, which accumulates errors, we observe that the distance between DIT and JIT gradually increases with each timestep. This indicates that DDIM inference gradually deviates from the original inversion trajectory under the influence of accumulated errors, causing a semantic drift between the inferred images and the original image. 

The three different forms of editing inference trajectories, although they end at different inference endpoints, are closer to JIT compared to DIT. This shows that despite modifying the image semantics according to the target prompt, they do not completely deviate from the semantic information of the original image. The involvement of the auxiliary chain during the inference process effectively constrains the semantics of the edited images. As can be seen in the right diagram, there is virtually no semantic gap between the editing trajectories and JIT in the initial stages of inference. This is due to the application of DCS, which ensures that the structural layout of the newly generated images remains the same as the original during the early stages of image generation, providing a foundation for maintaining high fidelity in the later stages of inference.

\section{Conclusions}
This paper presents ERDDCI, a new method designed to eliminate errors caused by local linearization assumptions under classifier-free guidance. This significantly improves the precision of image reconstruction and editing. ERDDCI uses a dual-chain inversion technology that eliminates the need for complex optimization procedures, making the process more time-efficient. Additionally, the DCS of ERDDCI enhances the precision and control in image editing tasks, which is particularl beneficial when dealing with large guidance scales, resulting in consistently reliable results.

\ifCLASSOPTIONcaptionsoff
\newpage
\fi

\bibliographystyle{./IEEEtran}
\bibliography{./ERDDCI}

% Generated by IEEEtran.bst, version: 1.12 (2007/01/11)
\begin{thebibliography}{10}
\providecommand{\url}[1]{#1}
\csname url@samestyle\endcsname
\providecommand{\newblock}{\relax}
\providecommand{\bibinfo}[2]{#2}
\providecommand{\BIBentrySTDinterwordspacing}{\spaceskip=0pt\relax}
\providecommand{\BIBentryALTinterwordstretchfactor}{4}
\providecommand{\BIBentryALTinterwordspacing}{\spaceskip=\fontdimen2\font plus
\BIBentryALTinterwordstretchfactor\fontdimen3\font minus \fontdimen4\font\relax}
\providecommand{\BIBforeignlanguage}[2]{{%
\expandafter\ifx\csname l@#1\endcsname\relax
\typeout{** WARNING: IEEEtran.bst: No hyphenation pattern has been}%
\typeout{** loaded for the language `#1'. Using the pattern for}%
\typeout{** the default language instead.}%
\else
\language=\csname l@#1\endcsname
\fi
#2}}
\providecommand{\BIBdecl}{\relax}
\BIBdecl

\bibitem{sohl2015deep}
J.~Sohl-Dickstein, E.~Weiss, N.~Maheswaranathan, and S.~Ganguli, ``Deep unsupervised learning using nonequilibrium thermodynamics,'' in \emph{International conference on machine learning}.\hskip 1em plus 0.5em minus 0.4em\relax PMLR, 2015, pp. 2256--2265.

\bibitem{ho2020denoising}
J.~Ho, A.~Jain, and P.~Abbeel, ``Denoising diffusion probabilistic models,'' \emph{Advances in neural information processing systems}, vol.~33, pp. 6840--6851, 2020.

\bibitem{song2020denoising}
J.~Song, C.~Meng, and S.~Ermon, ``Denoising diffusion implicit models,'' \emph{arXiv preprint arXiv:2010.02502}, 2020.

\bibitem{saharia2022palette}
C.~Saharia, W.~Chan, H.~Chang, C.~Lee, J.~Ho, T.~Salimans, D.~Fleet, and M.~Norouzi, ``Palette: Image-to-image diffusion models,'' in \emph{ACM SIGGRAPH 2022 conference proceedings}, 2022, pp. 1--10.

\bibitem{ho2022classifier}
J.~Ho and T.~Salimans, ``Classifier-free diffusion guidance,'' \emph{arXiv preprint arXiv:2207.12598}, 2022.

\bibitem{peebles2023scalable}
W.~Peebles and S.~Xie, ``Scalable diffusion models with transformers,'' in \emph{Proceedings of the IEEE/CVF International Conference on Computer Vision}, 2023, pp. 4195--4205.

\bibitem{chen2021multi}
X.~Chen, X.~Luo, J.~Weng, W.~Luo, H.~Li, and Q.~Tian, ``Multi-view gait image generation for cross-view gait recognition,'' \emph{IEEE Transactions on Image Processing}, vol.~30, pp. 3041--3055, 2021.

\bibitem{xu2023txt2img}
Y.~Xu, W.~Yu, P.~Ghamisi, M.~Kopp, and S.~Hochreiter, ``Txt2img-mhn: Remote sensing image generation from text using modern hopfield networks,'' \emph{IEEE Transactions on Image Processing}, 2023.

\bibitem{schwartz2018deepisp}
E.~Schwartz, R.~Giryes, and A.~M. Bronstein, ``Deepisp: Toward learning an end-to-end image processing pipeline,'' \emph{IEEE Transactions on Image Processing}, vol.~28, no.~2, pp. 912--923, 2018.

\bibitem{yang2023eliminating}
Z.~Yang, T.~Chu, X.~Lin, E.~Gao, D.~Liu, J.~Yang, and C.~Wang, ``Eliminating contextual prior bias for semantic image editing via dual-cycle diffusion,'' \emph{IEEE Transactions on Circuits and Systems for Video Technology}, vol.~34, no.~2, pp. 1316--1320, 2023.

\bibitem{zhao2021efficient}
Z.~Zhao, S.~Xu, J.~Zhang, C.~Liang, C.~Zhang, and J.~Liu, ``Efficient and model-based infrared and visible image fusion via algorithm unrolling,'' \emph{IEEE Transactions on Circuits and Systems for Video Technology}, vol.~32, no.~3, pp. 1186--1196, 2021.

\bibitem{rombach2022high}
R.~Rombach, A.~Blattmann, D.~Lorenz, P.~Esser, and B.~Ommer, ``High-resolution image synthesis with latent diffusion models,'' in \emph{Proceedings of the IEEE/CVF conference on computer vision and pattern recognition}, 2022, pp. 10\,684--10\,695.

\bibitem{ramesh2022hierarchical}
A.~Ramesh, P.~Dhariwal, A.~Nichol, C.~Chu, and M.~Chen, ``Hierarchical text-conditional image generation with clip latents,'' \emph{arXiv preprint arXiv:2204.06125}, vol.~1, no.~2, p.~3, 2022.

\bibitem{nichol2021glide}
A.~Nichol, P.~Dhariwal, A.~Ramesh, P.~Shyam, P.~Mishkin, B.~McGrew, I.~Sutskever, and M.~Chen, ``Glide: Towards photorealistic image generation and editing with text-guided diffusion models,'' \emph{arXiv preprint arXiv:2112.10741}, 2021.

\bibitem{gilboa2002forward}
G.~Gilboa, N.~Sochen, and Y.~Y. Zeevi, ``Forward-and-backward diffusion processes for adaptive image enhancement and denoising,'' \emph{IEEE transactions on image processing}, vol.~11, no.~7, pp. 689--703, 2002.

\bibitem{zhang2024real}
Y.~Zhang, J.~Xing, E.~Lo, and J.~Jia, ``Real-world image variation by aligning diffusion inversion chain,'' \emph{Advances in Neural Information Processing Systems}, vol.~36, 2024.

\bibitem{hertz2022prompt}
A.~Hertz, R.~Mokady, J.~Tenenbaum, K.~Aberman, Y.~Pritch, and D.~Cohen-Or, ``Prompt-to-prompt image editing with cross attention control,'' \emph{arXiv preprint arXiv:2208.01626}, 2022.

\bibitem{wallace2023edict}
B.~Wallace, A.~Gokul, and N.~Naik, ``Edict: Exact diffusion inversion via coupled transformations,'' in \emph{Proceedings of the IEEE/CVF Conference on Computer Vision and Pattern Recognition}, 2023, pp. 22\,532--22\,541.

\bibitem{pan2023effective}
Z.~Pan, R.~Gherardi, X.~Xie, and S.~Huang, ``Effective real image editing with accelerated iterative diffusion inversion,'' in \emph{Proceedings of the IEEE/CVF International Conference on Computer Vision}, 2023, pp. 15\,912--15\,921.

\bibitem{han2024proxedit}
L.~Han, S.~Wen, Q.~Chen, Z.~Zhang, K.~Song, M.~Ren, R.~Gao, A.~Stathopoulos, X.~He, Y.~Chen \emph{et~al.}, ``Proxedit: Improving tuning-free real image editing with proximal guidance,'' in \emph{Proceedings of the IEEE/CVF Winter Conference on Applications of Computer Vision}, 2024, pp. 4291--4301.

\bibitem{liu2022pseudo}
L.~Liu, Y.~Ren, Z.~Lin, and Z.~Zhao, ``Pseudo numerical methods for diffusion models on manifolds,'' \emph{arXiv preprint arXiv:2202.09778}, 2022.

\bibitem{mokady2023null}
R.~Mokady, A.~Hertz, K.~Aberman, Y.~Pritch, and D.~Cohen-Or, ``Null-text inversion for editing real images using guided diffusion models,'' in \emph{Proceedings of the IEEE/CVF Conference on Computer Vision and Pattern Recognition}, 2023, pp. 6038--6047.

\bibitem{biroli2024dynamical}
G.~Biroli, T.~Bonnaire, V.~de~Bortoli, and M.~Mézard, ``Dynamical regimes of diffusion models,'' 2024.

\bibitem{pehlivan2023styleres}
H.~Pehlivan, Y.~Dalva, and A.~Dundar, ``Styleres: Transforming the residuals for real image editing with stylegan,'' in \emph{Proceedings of the IEEE/CVF conference on computer vision and pattern recognition}, 2023, pp. 1828--1837.

\bibitem{brack2024ledits++}
M.~Brack, F.~Friedrich, K.~Kornmeier, L.~Tsaban, P.~Schramowski, K.~Kersting, and A.~Passos, ``Ledits++: Limitless image editing using text-to-image models,'' in \emph{Proceedings of the IEEE/CVF Conference on Computer Vision and Pattern Recognition}, 2024, pp. 8861--8870.

\bibitem{wu2022unifying}
C.~H. Wu and F.~De~la Torre, ``Unifying diffusion models' latent space, with applications to cyclediffusion and guidance,'' \emph{arXiv preprint arXiv:2210.05559}, 2022.

\bibitem{huberman2023edit}
I.~Huberman-Spiegelglas, V.~Kulikov, and T.~Michaeli, ``An edit friendly ddpm noise space: Inversion and manipulations,'' \emph{arXiv preprint arXiv:2304.06140}, 2023.

\bibitem{zhang2025exact}
G.~Zhang, J.~P. Lewis, and W.~B. Kleijn, ``Exact diffusion inversion via bidirectional integration approximation,'' in \emph{European Conference on Computer Vision}.\hskip 1em plus 0.5em minus 0.4em\relax Springer, 2025, pp. 19--36.

\bibitem{tang2024iterinv}
C.~Tang, K.~Wang, and J.~van~de Weijer, ``Iterinv: Iterative inversion for pixel-level t2i models,'' in \emph{2024 IEEE International Conference on Multimedia and Expo (ICME)}.\hskip 1em plus 0.5em minus 0.4em\relax IEEE, 2024, pp. 1--6.

\bibitem{huberman2024edit}
I.~Huberman-Spiegelglas, V.~Kulikov, and T.~Michaeli, ``An edit friendly ddpm noise space: Inversion and manipulations,'' in \emph{Proceedings of the IEEE/CVF Conference on Computer Vision and Pattern Recognition}, 2024, pp. 12\,469--12\,478.

\bibitem{miyake2023negative}
D.~Miyake, A.~Iohara, Y.~Saito, and T.~Tanaka, ``Negative-prompt inversion: Fast image inversion for editing with text-guided diffusion models,'' \emph{arXiv preprint arXiv:2305.16807}, 2023.

\bibitem{li2023stylediffusion}
S.~Li, J.~van~de Weijer, T.~Hu, F.~S. Khan, Q.~Hou, Y.~Wang, and J.~Yang, ``Stylediffusion: Prompt-embedding inversion for text-based editing,'' \emph{arXiv preprint arXiv:2303.15649}, 2023.

\bibitem{meng2021sdedit}
C.~Meng, Y.~He, Y.~Song, J.~Song, J.~Wu, J.-Y. Zhu, and S.~Ermon, ``Sdedit: Guided image synthesis and editing with stochastic differential equations,'' \emph{arXiv preprint arXiv:2108.01073}, 2021.

\bibitem{avrahami2022blended}
O.~Avrahami, D.~Lischinski, and O.~Fried, ``Blended diffusion for text-driven editing of natural images,'' in \emph{Proceedings of the IEEE/CVF Conference on Computer Vision and Pattern Recognition}, 2022, pp. 18\,208--18\,218.

\bibitem{couairon2022diffedit}
G.~Couairon, J.~Verbeek, H.~Schwenk, and M.~Cord, ``Diffedit: Diffusion-based semantic image editing with mask guidance,'' \emph{arXiv preprint arXiv:2210.11427}, 2022.

\bibitem{ruiz2023dreambooth}
N.~Ruiz, Y.~Li, V.~Jampani, Y.~Pritch, M.~Rubinstein, and K.~Aberman, ``Dreambooth: Fine tuning text-to-image diffusion models for subject-driven generation,'' in \emph{Proceedings of the IEEE/CVF Conference on Computer Vision and Pattern Recognition}, 2023, pp. 22\,500--22\,510.

\bibitem{kawar2023imagic}
B.~Kawar, S.~Zada, O.~Lang, O.~Tov, H.~Chang, T.~Dekel, I.~Mosseri, and M.~Irani, ``Imagic: Text-based real image editing with diffusion models,'' in \emph{Proceedings of the IEEE/CVF Conference on Computer Vision and Pattern Recognition}, 2023, pp. 6007--6017.

\bibitem{su2022dual}
X.~Su, J.~Song, C.~Meng, and S.~Ermon, ``Dual diffusion implicit bridges for image-to-image translation,'' \emph{arXiv preprint arXiv:2203.08382}, 2022.

\bibitem{kim2022diffusionclip}
G.~Kim, T.~Kwon, and J.~C. Ye, ``Diffusionclip: Text-guided diffusion models for robust image manipulation,'' in \emph{Proceedings of the IEEE/CVF Conference on Computer Vision and Pattern Recognition}, 2022, pp. 2426--2435.

\bibitem{zhu2023boundary}
Y.~Zhu, Y.~Wu, Z.~Deng, O.~Russakovsky, and Y.~Yan, ``Boundary guided learning-free semantic control with diffusion models,'' in \emph{Thirty-seventh Conference on Neural Information Processing Systems}, 2023.

\bibitem{bar2022text2live}
O.~Bar-Tal, D.~Ofri-Amar, R.~Fridman, Y.~Kasten, and T.~Dekel, ``Text2live: Text-driven layered image and video editing,'' in \emph{European conference on computer vision}.\hskip 1em plus 0.5em minus 0.4em\relax Springer, 2022, pp. 707--723.

\bibitem{sadat2024eliminating}
S.~Sadat, O.~Hilliges, and R.~M. Weber, ``Eliminating oversaturation and artifacts of high guidance scales in diffusion models,'' \emph{arXiv preprint arXiv:2410.02416}, 2024.

\bibitem{lin2014microsoft}
T.-Y. Lin, M.~Maire, S.~Belongie, J.~Hays, P.~Perona, D.~Ramanan, P.~Doll{\'a}r, and C.~L. Zitnick, ``Microsoft coco: Common objects in context,'' in \emph{Computer Vision--ECCV 2014: 13th European Conference, Zurich, Switzerland, September 6-12, 2014, Proceedings, Part V 13}.\hskip 1em plus 0.5em minus 0.4em\relax Springer, 2014, pp. 740--755.

\end{thebibliography}

\end{document}